\documentclass[10pt,twocolumn,letterpaper]{article}

\usepackage{cvpr}
\usepackage{times}
\usepackage{epsfig}
\usepackage{graphicx}
\usepackage{amsmath}
\usepackage{amssymb}

\usepackage{adjustbox}

\usepackage{booktabs}

\usepackage{array}

\newcommand{\ditto}{$\cdot$}  

\usepackage[table]{xcolor}
\newcommand{\maxf}[1]{{\cellcolor[gray]{0.8}} #1}

\usepackage[group-separator={,}]{siunitx}  
\usepackage[acronym]{glossaries}
\newacronym{ar}{AR}{augmented reality}
\newacronym{bn}{BN}{batch normalization}
\newacronym{dprime}{$d'$}{decidability index}
\newacronym{eer}{EER}{equal error rate}
\newacronym{ekyt}{EKYT}{Eye Know You Too}
\newacronym{et}{ET}{eye tracking}
\newacronym{far}{FAR}{false acceptance rate}
\newacronym{frr}{FRR}{false rejection rate}
\newacronym{icc}{ICC}{intraclass correlation coefficient}
\newacronym{ir}{IR}{identification rate}
\newacronym{ms}{MS}{multi-similarity}
\newacronym{relu}{ReLU}{rectified linear unit}
\newacronym{roc}{ROC}{receiver operating characteristic}
\newacronym{sd}{SD}{standard deviation}
\newacronym{vr}{VR}{virtual reality}
\newacronym{xr}{XR}{extended reality}

\newcommand{\expBothBothFolds}{18}
\newcommand{\expBothBothBigFolds}{6}
\newcommand{\expBothBothFoldsFour}{15}

\newcommand{\expBothBothBig}{5}
\newcommand{\expLeftVisual}{1}

\newcommand{\expBothVisual}{2}

\newcommand{\expBothBoth}{3}
\newcommand{\expBothBothHigh}{16}
\newcommand{\expBothBothLow}{17}
\newcommand{\expBothBothQuarter}{7}
\newcommand{\expBothBothHalf}{8}
\newcommand{\expBothBothOne}{9}
\newcommand{\expBothBothTwo}{10}
\newcommand{\expBothBothThree}{11}
\newcommand{\expBothBothFour}{12}
\newcommand{\expBothBothFive}{13}
\newcommand{\expBothBothSix}{14}
\newcommand{\expBothDiff}{4}

\usepackage[pagebackref=true,breaklinks=true,colorlinks,bookmarks=false]{hyperref}

\begin{document}

\title{Establishing a Baseline for Gaze-driven Authentication Performance in VR: A Breadth-First Investigation on a Very Large Dataset}

\author{Dillon Lohr$^{1,2}$, Michael J. Proulx$^{2}$, Oleg Komogortsev$^{1,2}$\\
$^{1}$Texas State University, San Marcos, TX, USA $^{2}$Meta Reality Labs Research, Redmond, WA, USA\\
{\tt\small djl70@txstate.edu, michaelproulx@meta.com, ok@txstate.edu}
}

\maketitle
\thispagestyle{empty}

\begin{abstract}
This paper performs the crucial work of establishing a baseline for gaze-driven authentication performance to begin answering fundamental research questions using a very large dataset of gaze recordings from \num{9202}~people with a level of \gls{et} signal quality equivalent to modern consumer-facing \gls{vr} platforms.
The size of the employed dataset is at least an order-of-magnitude larger than any other dataset from previous related work.
Binocular estimates of the optical and visual axes of the eyes and a minimum duration for enrollment and verification are required for our model to achieve a \gls{frr} of below \num{3}\% at a \gls{far} of \num{1} in \num{50000}.
In terms of identification accuracy which decreases with gallery size, we estimate that our model would fall below chance-level accuracy for gallery sizes of \num{148000} or more.
Our major findings indicate that gaze authentication can be as accurate as required by the FIDO standard when driven by a state-of-the-art machine learning architecture and a sufficiently large training dataset.
\end{abstract}
\glsresetall

\section{Introduction}
Accurate, spoof-resistant, and privacy-aware user authentication on \gls{xr} devices is an important topic of research necessary to provide the highest levels of security to users of such devices.
Passwords can be difficult to remember and inconvenient to enter, and they are not suitable for continuous authentication.
Comparatively, eye gaze authentication~\cite{Kasprowski2004} might reduce cognitive load and increase convenience.
For the best user experience and highest security, gaze signatures would need to exhibit high distinctiveness, permanence, collectability, and universality~\cite{jain2004introduction}, and they would also need to be highly spoof-resistant and enable fast accept/reject decisions.

Eye gaze authentication can leverage existing \gls{et} hardware in modern \gls{xr} devices to provide accurate and spoof-resistant unlocking of such devices.
It is also important to state that while devices such as the Apple Vision Pro~\cite{AppleVisionPro} employ fine-tuned hardware for user authentication via iris patterns, iris-based authentication is not always possible for gaze estimation pipelines that employ non-image sensors~\cite{AdHawk,Inseye,golard_2021_ultrasound,stoffregen_2022_eventsensor,whitmire_2016_eyecontact} or for image-based gaze estimation pipelines with optics and algorithms that are fine-tuned to increase \gls{et} signal quality and reduce power consumption at the expense of the visibility of fine-grained information related to iris ridges~\cite{EyeLink1000,palmero_2021_openeds2020}.
Another advantage of gaze authentication is its high spoof-resistance~\cite{makowski2021deepeyedentificationlive,raju2022irisprintattack,Rigas2015} due to incorporating not only static parameters of gaze such as instantaneous measurement of a person's optical and visual axes but also the dynamics of gaze which includes the variety and complexity of various eye movement types~\cite{leigh2006neurology}.

In this work, we report on various aspects of user authentication using the \textit{GazePro} dataset.
GazePro contains \gls{et} signals from a population of \num{9202}~people recorded at \num{72}~Hz using a \gls{vr} headset that provides a level of \gls{et} signal quality equivalent to a Meta Quest Pro~\cite{aziz_2024_evaluation,Wei2023}.
Our use of such hardware is unlike many previous studies~\cite{lohr_2022_ekyt,makowski2021deepeyedentificationlive} which often utilize laboratory-grade \gls{et} systems with signal quality characteristics that are currently unachievable in consumer-grade \gls{xr} devices.

We report a best-case result of \num{2.4}\% \gls{frr} at a \num{1}-in-\num{50000} \gls{far} when using \num{20}~seconds of gaze data during a random saccade (jumping dot) task.
To the best of our knowledge, this is the first time such a high level of authentication accuracy has been achieved via gaze alone.
Thus, we are the first to demonstrate that gaze can potentially provide meaningful user authentication performance under FIDO guidelines~\cite{FIDO2024}.
Importantly, this result was achieved using the current state-of-the-art machine learning architecture, \gls{ekyt}~\cite{lohr_2022_ekyt}, signifying the importance of large-scale datasets for achieving meaningful performance at the level of \gls{et} signal quality employed in our work.
It is also important to note that identification rates decrease with increasing gallery sizes, and we roughly estimate that we might reach random-chance identification rates starting at a gallery size of about \num{148000}~identities.

Here we perform a breadth-first investigation of several fundamental research questions to provide necessary baselines for future work in this important research area and to begin answering several of these questions for the first time:
(RQ1)~Does the use of binocular gaze improve authentication accuracy over monocular gaze?
(RQ2)~Does the use of optical axis estimates in addition to visual axis estimates improve authentication accuracy?
(RQ3)~What aspects of training the state-of-the-art machine learning architecture result in the best authentication accuracy?
(RQ4)~How does the size of the training dataset affect authentication accuracy?
(RQ5)~How quickly does identification accuracy degrade as the gallery size increases?
(RQ6)~How does \gls{et} signal quality, specifically spatial accuracy, affect authentication accuracy?
(RQ7)~How does the length of the employed gaze signal affect authentication accuracy?
(RQ8)~Can authentication accuracy be improved by increasing the duration of the enrollment session?
(RQ9)~Can gaze authentication be task-independent?
(RQ10)~Is the permanence of the learned embedding features correlated with authentication accuracy?

\section{Background}
Since the seminal work of Kasprowski \& Ober~\cite{Kasprowski2004}, gaze-based authentication has been studied extensively~\cite{Abdelwahab2019,Bargary2017,Eberz2019,Friedman2017,Galdi2016,Jager2020,Jia2018,Kasprowski2018,Katsini2020,Li2018,lohr_2022_ekyt,makowski2021deepeyedentificationlive,Rigas2017,Zhang2017}.
End-to-end deep learning approaches have only recently been explored~\cite{Abdelwahab2019,Jager2020,Jia2018,lohr_2022_ekyt,makowski2021deepeyedentificationlive}, with the current state-of-the-art model being \gls{ekyt}~\cite{lohr_2022_ekyt}.
The \gls{ekyt} architecture and training methodology are employed for all experiments in the present study.
Some modifications to the original training parameters are also explored in pursuit of even better performance on the employed dataset.

Studies in this domain typically employed either monocular gaze data (such as~\cite{lohr_2022_ekyt}) or binocular gaze data (such as~\cite{makowski2021deepeyedentificationlive}), but the literature currently lacks studies that compare performance with monocular versus binocular data.
Binocular data contains information that cannot be measured from monocular data alone, such as eye dominance and vergence angle.
Such additional information might enable increased authentication accuracy using binocular data compared to monocular data.
The present study provides such a performance comparison.

\begin{figure}[t]
    \centering
    \includegraphics[scale=1.8,trim={0 0.35cm 0 0.35cm},clip]{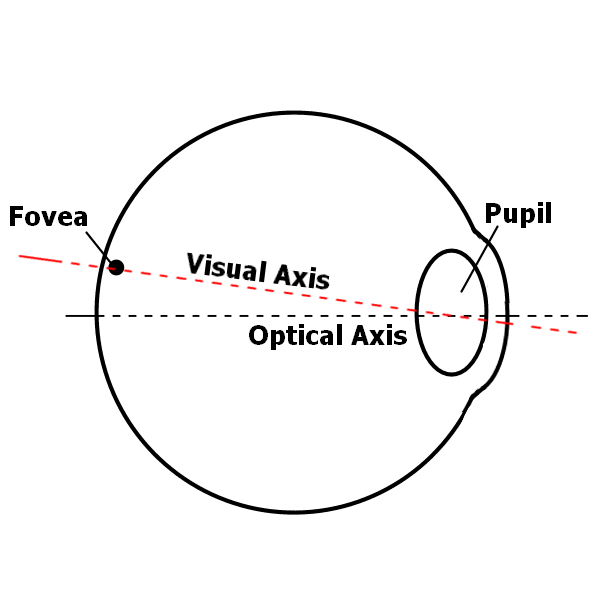}
    \caption{A simplified model of the eye, visualizing the imaginary optical and visual axes.  The optical axis passes through the eyeball center and the pupil center.  The visual axis connects the center of the fovea (``foveola'') and the gaze target (``object of regard'').}
    \label{fig:eye-model}
\end{figure}

The vast majority of studies in this domain employed estimates of the visual axis (sometimes called the line of sight) to represent gaze.
With current \gls{et} technology which cannot directly measure the position of the fovea, the visual axis can be estimated only after performing a user-specific calibration procedure.
But there are scenarios such as continuous authentication in which it is impractical to precede every verification attempt with a calibration procedure.
In contrast, the optical axis (sometimes called the pupillary axis) can be estimated without a user-specific calibration procedure~\cite{Barsingerhorn2018}.
The optical and visual axes are illustrated in Figure~\ref{fig:eye-model}.
Even in scenarios where the visual axis can be estimated, perhaps a combination of both the optical and visual axes might improve authentication accuracy.
To the best of our knowledge, the present study is the first to assess gaze-based authentication when using the optical axis and when using a combination of both the optical and visual axes.

Currently, the largest public \gls{et} datasets used for gaze authentication contain on the order of hundreds of people (e.g., \num{150}~in JuDo1000~\cite{Makowski2020}, \num{322}~in GazeBase~\cite{griffith2021gazebase}, and \num{407}~in GazeBaseVR~\cite{lohr_2023_gazebasevr}).
In contrast, the present study employs GazePro, a dataset of \num{9202}~people, which is an order-of-magnitude larger than existing datasets.
The GazePro dataset affords the present study the unique opportunity to explore the applicability of reliable gaze authentication on a never-before-seen scale.

There are several characteristics commonly used to describe \gls{et} signal quality~\cite{holmqvist2011comprehensive}, such as sampling rate, spatial accuracy (error), and spatial precision (jitter).
Some studies have investigated how different sampling rates and levels of spatial precision affect authentication performance~\cite{lohr_2022_ekyt,prasse2020relationship}, with the general consensus being that lower sampling rates and/or worse spatial precision lead to worse authentication performance.
There is also evidence~\cite{raju_2023_noise} that high-frequency (``noise'') components of gaze signals carry person-specific information with features that are reliable across short test-retest intervals.
It is currently hypothesized that the gaze estimation pipeline creates person-specific noise.
But the effect of spatial accuracy on authentication performance is underexplored.
A recent study by Raju et al.~\cite{raju_2024_embeddings} suggested that authentication performance does vary across different levels of spatial accuracy.
Surprisingly, that study reported that they achieved \textit{better} authentication performance on a subset of users from GazeBaseVR~\cite{lohr_2023_gazebasevr} with \textit{worse} spatial accuracy than on those with better spatial accuracy.
The present study further explores the effect of spatial accuracy on authentication performance across a larger user population.

Test-set population sizes vary across studies in this domain, from \num{25}~\cite{makowski2021deepeyedentificationlive} to \num{59}~\cite{lohr_2022_ekyt} to \num{149}~\cite{Friedman2017}.
Few studies meet the FIDO guidelines~\cite{FIDO2024} of a test-set population size of \num{245}~users.
Further, the theoretical analysis by Friedman et al.~\cite{Friedman2020} used synthetic feature distributions to show that verification accuracy remained stable across population sizes while identification accuracy decreased with larger population sizes.
Until now, no study has had a large enough dataset size to confirm these findings in practice in the \gls{et} domain.

Similarly, the minimum duration of each verification attempt also widely varied across studies, from \num{1}~second~\cite{makowski2021deepeyedentificationlive} to \num{5}~seconds~\cite{lohr_2022_ekyt} to \num{60}~seconds~\cite{Friedman2017}.
As one might expect, studies have found that authentication accuracy generally improved as more data was used for each verification attempt~\cite{lohr_2022_ekyt,makowski2021deepeyedentificationlive}.
The study by Lohr \& Komogortsev~\cite{lohr_2022_ekyt} used the same amount of data for both enrollment and verification and found that at least \num{30}~seconds of gaze data was necessary to achieve their best performance of \num{5.08}\% \gls{frr} at a \num{1}-in-\num{10000} \gls{far}.
The study by Makowski et al.~\cite{makowski2021deepeyedentificationlive} instead used a fixed enrollment duration of \num{24}~seconds and only varied the verification duration.
The present study further explores how authentication accuracy is affected by data length across a much larger population of people.

Most studies in this domain trained and evaluated on the same task, such as a reading task (e.g.,~\cite{Friedman2017}) or a random saccade (jumping dot) task (e.g.,~\cite{makowski2021deepeyedentificationlive}).
A study by Holland \& Komogortsev~\cite{holland2013stimulus} was one of the first to demonstrate that similar gaze authentication performance could be achieved on different tasks.
Lohr \& Komogortsev~\cite{lohr_2022_ekyt} later demonstrated that a model trained across a variety of different tasks could achieve similar authentication performance across all such tasks, including an additional task that was not part of the training set.
The present study further explores the task-independence of gaze authentication across a larger user population.

Friedman et al.~\cite{Friedman2020b} have previously demonstrated that high temporal persistence (permanence) of learned features is important for improved authentication performance.
Raju et al.~\cite{raju_2024_embeddings} have also found a high correlation between permanence and authentication performance.
Lohr \& Komogortsev~\cite{lohr_2022_ekyt} have shown that gaze signatures can exhibit high permanence over periods as long as \num{3}~years.
The present study provides further insight into the relationship between permanence (on a short test-retest interval) and authentication performance on a dataset with a much larger user population.

\section{Methodology}
An overview of our methodology is presented here and summarized by the diagram in Figure~\ref{fig:overview-diagram}.
More details can be found in the supplementary material for this manuscript.
\begin{figure*}
    \centering
    \includegraphics[width=\linewidth]{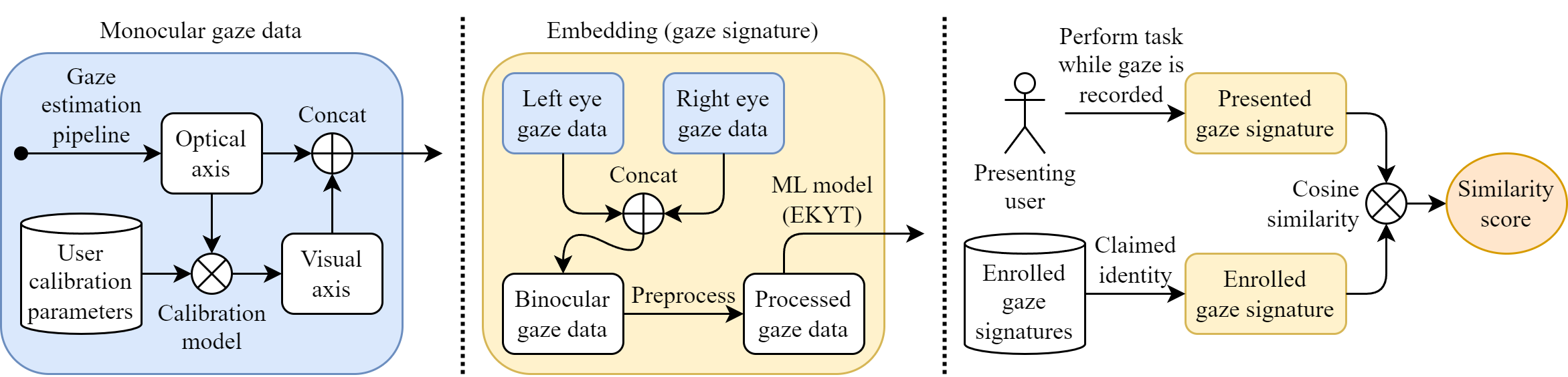}
    \caption{Overview diagram of the methodology for computing similarity scores.}
    \label{fig:overview-diagram}
\end{figure*}

\subsection{Dataset}
\label{sec:dataset}
We employed \textit{GazePro}, an internal dataset of gaze signals collected from \num{9202}~participants which is an order-of-magnitude more participants than have been used for studies in this domain (e.g.,~\cite{griffith2021gazebase,lohr_2023_gazebasevr,Makowski2020}).
Having access to such a large population gives us the unique opportunity to evaluate gaze authentication at a previously unattainable scale.
The dataset is split into a training set of \num{6747}~users and a testing set of \num{2455}~users.

Of the \num{9202}~participants in GazePro, \num{3673}~self-reported as male, \num{5376}~as female, and \num{153}~as neither male nor female.
The participants' ages ranged from \num{13}~to \num{88}~years (median=\num{33}, IQR=\num{18}).
All participants had normal or corrected-to-normal vision.
Data collection went through the appropriate IRB approval protocol and received appropriate consent from participants.
Additional details about how the dataset was collected and participant characteristics can be found in the supplementary material.

Gaze data in GazePro were collected at \num{72}~Hz on a \gls{vr} headset with \gls{et} signal quality and other general characteristics equivalent to a Meta Quest Pro~\cite{aziz_2024_evaluation,Wei2023}.
Each recording contains both optical and visual axis estimates from both the left and right eyes.
The visual axis is a linear transformation of the optical axis.
The parameters of the linear model (essentially, the \textit{angle kappa} between the optical and visual axes) are fit on a per-user basis via a short calibration procedure.

\subsection{Model training}
\label{sec:model-training}
We closely followed the same training methodology that was used for \gls{ekyt}~\cite{lohr_2022_ekyt} with only a couple of differences.
Models were re-implemented using PyTorch and trained on an NVIDIA V100 GPU with 16~GB VRAM.

We employed the current state-of-the-art network architecture for gaze authentication, \gls{ekyt}~\cite{lohr_2022_ekyt}.
The network transforms a gaze sequence in $\mathbb{R}^{\text{C} \times \text{T}}$ to an embedding in $\mathbb{R}^{128}$, where $\text{C} = 2 \times (\text{number of eyes}) \times (\text{number of axes})$ is the number of input channels and $\text{T} = 360$ is the number of time steps in the input sequence (\num{5}~seconds at \num{72}~Hz).
The input channels are horizontal and vertical velocity components, taken from either the left or right eye (or both) and from either the optical or visual axis (or both).

The model is trained using \gls{ms} loss~\cite{Wang2019}, which encourages the model to learn a well-clustered embedding space with smaller distances between embeddings from the same user and larger distances between embeddings from different users.
We note that Lohr \& Komogortsev~\cite{lohr_2022_ekyt} included an additional classification loss term, but we exclude that term along with the associated classification layer.
This was done because the original work showed that training with \gls{ms} loss alone slightly improved authentication performance in some instances.

Although the original approach by Lohr \& Komogortsev~\cite{lohr_2022_ekyt} trained \num{4}~models using a \num{4}-fold split of the training set, in most of our experiments we instead train a single model on the full training set.
We found that training a single model (still using an embedding size of~\num{128}) generally performs better on our testing set than the original approach that concatenates the embeddings from \num{4}~models.

\subsection{Model evaluation}
\label{sec:model-evaluation}
We consider two main evaluation scenarios.
In both scenarios, the embedding of a gaze sequence is obtained by first embedding each non-overlapping segment of \num{5}~seconds and then taking the mean across all such segment embeddings.

The first scenario is \textit{verification}.
A presenting user claims to be a specific person (e.g., the owner of a \gls{vr} headset), their gaze signature is captured, and an accept/reject decision must be made based on the similarity of the gaze signatures captured from the person with the claimed identity at enrollment time (e.g., during account creation) and from the presenting user at verification time.
We measure performance in this scenario using \gls{eer} and \gls{frr} at \num{1}-in-\num{50000} \gls{far} (denoted $\text{FRR}_{0.002\%}$).

The second scenario is \textit{identification}.
A user's gaze signature is captured, the similarity scores are computed between that user's gaze signature and all gaze signatures from the population of enrolled participants, and the participant with the highest similarity is determined to be the best match.
It is possible to reject a user if the highest similarity score is below some predetermined threshold, in which case it is assumed the user is not part of the enrolled population; but in our case, we evaluated identification rates only on users who were part of the enrolled population.
We measure performance in this scenario using Rank-1 \gls{ir}.

\section{Results}
\begin{table*}
    \centering
    \caption{Verification performance for a selection of experiments of interest.  Lower values indicate better performance.  The best results are highlighted in gray.  To reduce visual clutter, the symbol ``\ditto'' is used to indicate that the value is repeated from the row above.  In the ``Eye'' column, ``L'' is the left eye and ``R'' is the right eye.  In the ``Axis'' column, ``O'' is the optical axis and ``V'' is the visual axis.}
    \label{tab:highlighted-results}
    \begin{tabular}{l l l l l S[table-format=2.2] S[table-format=3.1]}
        \toprule
        Experiment & Eye & Axis & Train set & Test set & {EER (\%)} & {$\text{FRR}_{0.002\%}$ (\%)} \\
        \midrule
        \expLeftVisual{} & L & V & Full & Full & 8.53 & 99.2 \\
        \expBothVisual{} & L,R	& {\ditto}	& {\ditto}	& {\ditto} & 3.73	& 93.1 \\
        \midrule
        \expBothBoth{} & L,R	& O,V	& Full	& Full & 0.08	& 17.7 \\
        \expBothDiff{} & {\ditto}	& $\text{V} - \text{O}$	& {\ditto}	& {\ditto} & 0.27	& 45.6 \\
        \midrule
        \expBothBothBig{} & L,R & O,V & Full, \num{759} epochs, $m=1024$ & Full & \maxf{0.04} & 4.1 \\
        \expBothBothBigFolds{} & {\ditto} & {\ditto} & \num{4} folds, \num{1000} epochs, $m=1024$ & {\ditto} & \maxf{0.04} & \maxf{2.4} \\
        \midrule
        \expBothBothQuarter{} & L,R	& O,V	& N=\hphantom{0}250	& Full & 1.59	& 92.2 \\
        \expBothBothHalf{} & {\ditto}	& {\ditto}	& N=\hphantom{0}500	& {\ditto} & 0.98	& 84.1 \\
        \expBothBothOne{} & {\ditto}	& {\ditto}	& N=1000	& {\ditto}	& 0.66	& 82.6 \\
        \expBothBothTwo{} & {\ditto}	& {\ditto}	& N=2000	& {\ditto}	& 0.61	& 78.9 \\
        \expBothBothThree{} & {\ditto}	& {\ditto}	& N=3000	& {\ditto}	& 0.10	& 22.0 \\
        \expBothBothFour{} & {\ditto}	& {\ditto}	& N=4000	& {\ditto}	& 0.42	& 71.7 \\
        \expBothBothFive{} & {\ditto}	& {\ditto}	& N=5000	& {\ditto}	& 0.07	& 13.7 \\
        \expBothBothSix{} & {\ditto}	& {\ditto}	& N=6000	& {\ditto}	& 0.06	& 12.6 \\
        \midrule
        \expBothBothFoldsFour{}@\num{250} & L,R	& O,V	& \num{4}~folds, N=4000	& N=\hphantom{0}250 & 0.42	& 65.9 \\
        \expBothBothFoldsFour{}@\num{500} & {\ditto}	& {\ditto}	& {\ditto}	& N=\hphantom{0}500	& 0.45	& 68.9 \\
        \expBothBothFoldsFour{}@\num{1}K & {\ditto}	& {\ditto}	& {\ditto}	& N=1000	& 0.41	& 70.4 \\
        \expBothBothFoldsFour{}@\num{2}K & {\ditto}	& {\ditto}	& {\ditto}	& N=2000	& 0.43	& 69.9 \\
        \expBothBothFoldsFour{}@\num{3}K & {\ditto}	& {\ditto}	& {\ditto}	& N=3000	& 0.44	& 71.3 \\
        \expBothBothFoldsFour{}@\num{4}K & {\ditto}	& {\ditto}	& {\ditto}	& N=4000	& 0.44	& 71.0 \\
        \expBothBothFoldsFour{}@\num{5}K & {\ditto}	& {\ditto}	& {\ditto}	& N=5000	& 0.44	& 70.9 \\
        \midrule
        \expBothBothHigh{}@HighErr & L,R	& O,V	& High err.	& High err. & 1.00	& 75.9 \\
        \expBothBothHigh{}@MidErr & {\ditto}	& {\ditto}	& {\ditto}	& Mid err. & 0.41	& 75.9 \\
        \expBothBothHigh{}@LowErr & {\ditto}	& {\ditto}	& {\ditto}	& Low err. & 0.73	& 76.7 \\
        \expBothBothLow{}@HighErr & {\ditto}	& {\ditto}	& Low err.	& High err. & 2.25	& 91.8 \\
        \expBothBothLow{}@MidErr & {\ditto}	& {\ditto}	& {\ditto}	& Mid err. & 0.41	& 70.6 \\
        \expBothBothLow{}@LowErr & {\ditto}	& {\ditto}	& {\ditto}	& Low err. & 0.42	& 92.2 \\
        \midrule
        \expBothBothBig{}@5s & L,R	& O,V	& Full, \num{759} epochs, $m=1024$	& Full & 0.18	& 41.9 \\
        \expBothBothBig{}@10s & {\ditto}	& {\ditto}	& {\ditto}	& {\ditto} & 0.08	& 20.6 \\
        \expBothBothBig{}@15s & {\ditto}	& {\ditto}	& {\ditto}	& {\ditto} & 0.05	& 8.8 \\
        \midrule
        \expBothBothFolds{}@SAC & L,R & O,V & \num{4} folds & Full & 0.10 & 22.9 \\
        \expBothBothFolds{}@PUR & {\ditto} & {\ditto} & {\ditto} & {\ditto} & 0.10 & 24.8 \\
        \bottomrule
    \end{tabular}
\end{table*}

We highlight the verification performance under various conditions in Table~\ref{tab:highlighted-results}.
Unless otherwise noted, results were computed during a random saccade (jumping dot) task with the full training set (\num{6747}~users) and the full testing set (\num{2455}~users), using a model trained for \num{100}~epochs with a minibatch size of $m=256$~samples, and using \num{20}~seconds of gaze data for both enrollment and verification.
Not all \num{2455}~users in the testing set were recorded for one or both of the tasks used for enrollment/verification, so we had fewer than \num{2455}~genuine matches.
Instead, in this case we obtained \num{2359}~genuine matches and \num{5707331}~impostor matches from a matrix of $2385 \times 2394$~match scores.

We assigned each experiment a unique ID to enable easier reference.
Additional experiments can be found in the supplementary material.

Lohr \& Komogortsev~\cite{lohr_2022_ekyt} used measures of the left eye's visual axis.
As a baseline, we did the same in Experiment~\expLeftVisual{} and achieved an \gls{eer} of \num{8.53}\% and a $\text{FRR}_{0.002\%}$ of \num{99.2}\%.

\paragraph{(RQ1) Monocular gaze vs binocular gaze.}
Although our baseline experiment used data from a single eye, many \gls{et}-capable \gls{xr} systems, including the one employed for GazePro, track both eyes.
Additional information, such as eye dominance and vergence angle, is encoded in binocular data.
For that reason, authentication accuracy might be increased due to this additional information.

Experiment~\expBothVisual{}, which used the visual axis from both the left and right eyes, can be compared against our monocular baseline (Experiment~\expLeftVisual{}).
This experiment achieved an \gls{eer} of \num{3.73}\% and a $\text{FRR}_{0.002\%}$ of \num{93.1}\%.
These results were substantially better than the monocular baseline.

\paragraph{(RQ2) Optical axis and visual axis.}
The majority of studies in this domain use estimates of the visual axis.
Working under the condition that we have access to a user's calibration parameters to estimate the visual axis, what if both the optical and visual axes were fed into the model?
This would implicitly provide information to the model about the user's calibration parameters, so we would expect better performance using both axes than visual axis alone.

Experiment~\expBothBoth{}, which used both the optical and visual axes from both the left and right eyes, can be compared against our visual axis baseline (Experiment~\expBothVisual{}).
This experiment achieved an \gls{eer} of \num{0.08}\% and a $\text{FRR}_{0.002\%}$ of \num{17.7}\%.
These results were substantially better than the visual axis baseline.

To further demonstrate the effectiveness of the user-specific calibration parameters, Experiment~\expBothDiff{} uses the difference of the two axes (visual minus optical) for both the left and right eyes.
This experiment achieves an \gls{eer} of \num{0.27}\% and a $\text{FRR}_{0.002\%}$ of \num{45.6}\%.
These results were still substantially better than the visual axis baseline but worse than when using both axes together.

\paragraph{(RQ3) Training epochs and minibatch size.}
We found that increasing the number of training epochs from~\num{100} to~\num{1000} yielded a large improvement in performance.
Simultaneously increasing the minibatch size from $m=256$ to $m=1024$ further improved performance.

\begin{figure}
    \centering
    \includegraphics[width=0.49\linewidth]{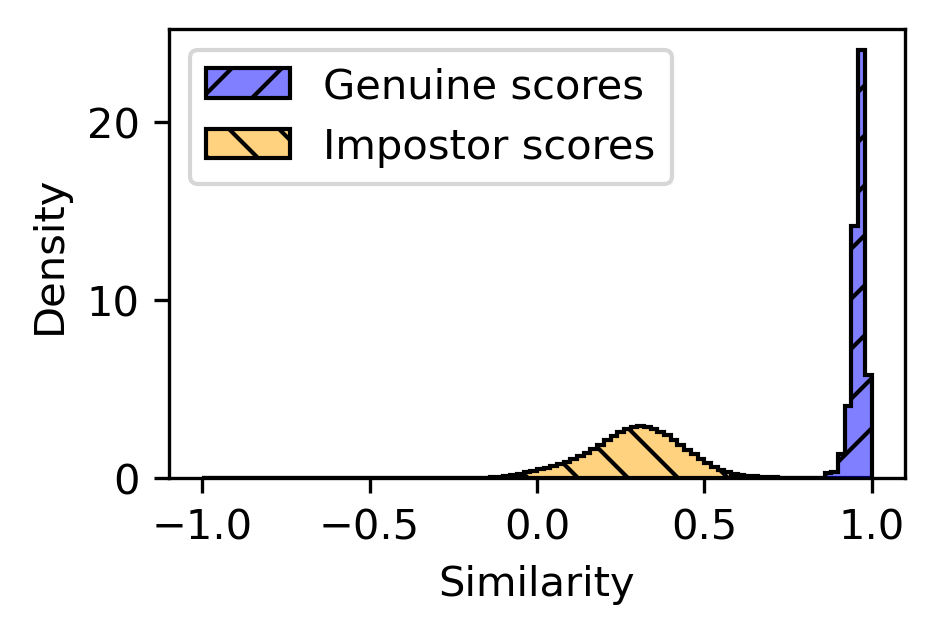}
    \includegraphics[width=0.49\linewidth]{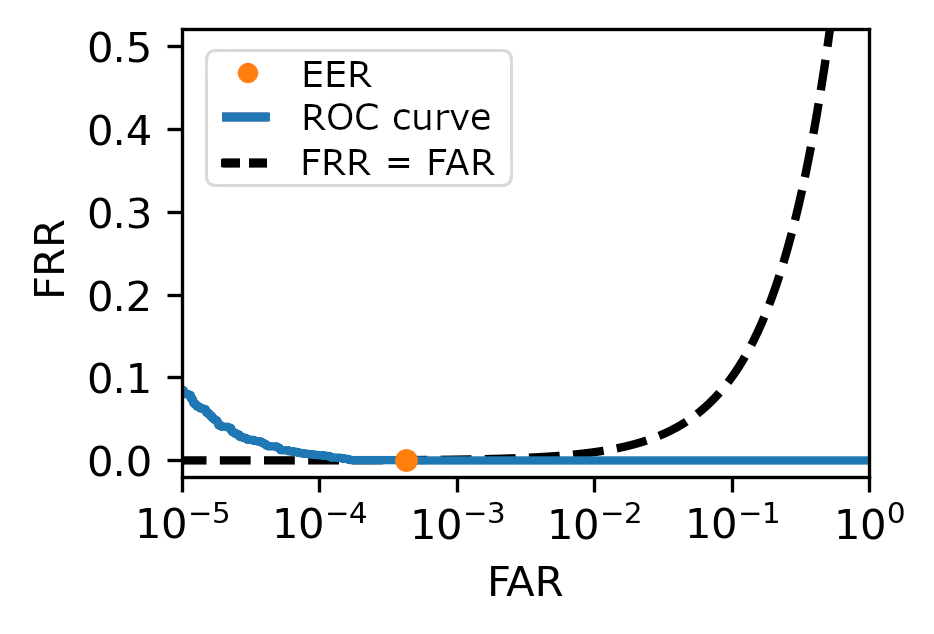}
    \caption{Qualitative performance from Experiment~\expBothBothBig{}.  The left figure shows histograms of the genuine and impostor similarity score distributions.  The right figure shows the \gls{roc} curve.}
    \label{fig:sim-roc-exp54}
\end{figure}

In Experiment~\expBothBothBig{}, we achieved \num{0.04}\% \gls{eer} and \num{4.1}\% $\text{FRR}_{0.002\%}$ when training for \num{759}~epochs (terminated early due to time constraints) with a minibatch size of $m=1024$, and when using both the optical and visual axes from both the left and right eyes.
These results were substantially better than Experiment~\expBothBoth{} which trained for \num{100}~epochs with a minibatch size of $m=256$.
The genuine and impostor similarity distributions and \gls{roc} curve for this experiment are presented in Figure~\ref{fig:sim-roc-exp54}.

Our best results were from Experiment~\expBothBothBigFolds{} where we achieved \num{0.04}\% \gls{eer} and \num{2.4}\% $\text{FRR}_{0.002\%}$ using \num{20}~seconds of a random saccade task for both enrollment and verification.
To obtain these results, we trained an ensemble of \num{4}~models using a \num{4}-fold split of the training set similar to~\cite{lohr_2022_ekyt}.
Each model was trained for \num{1000}~epochs with a minibatch size of $m=1024$ and used both the optical and visual axes from both the left and right eyes.

\paragraph{(RQ4) Training population size.}
Considering that the GazePro dataset has an order-of-magnitude more participants than the largest public \gls{et} datasets such as GazeBase~\cite{griffith2021gazebase} and GazeBaseVR~\cite{lohr_2023_gazebasevr}, we were curious how important it was to have such a large dataset.
For these experiments, we trained the model on a smaller subset of subjects from the training set to see how performance changed as the amount of training data increased.
The test set is unmodified to facilitate comparisons across experiments.

\begin{figure}
    \centering
    \includegraphics[width=0.9\linewidth]{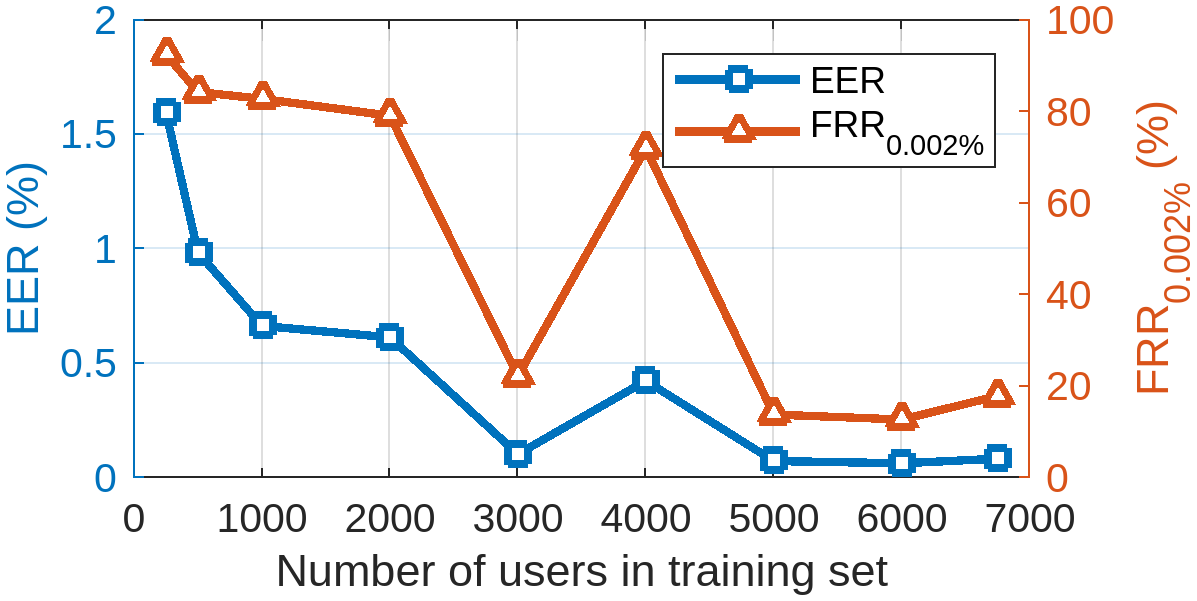}
    \caption{Authentication performance vs training population size.  Measurements are from Experiments~\expBothBoth{} and \expBothBothQuarter{}--\expBothBothSix{}.}
    \label{fig:perf-vs-train-size}
\end{figure}

We observed, perhaps not surprisingly, that performance generally continued to improve as the training population increased from \num{250} to \num{6000} users (see Figure~\ref{fig:perf-vs-train-size} for a visualization).
The improvement is not necessarily monotonic (see for example Experiments~\expBothBoth{} and~\expBothBothFour{}), which may be due to the inherent randomness of training and minibatch sampling leading to some models happening to settle on poor local optima.
We see the largest improvement when increasing the training population from \num{2000} to \num{3000} users, but it is difficult to determine whether this was just the first time the model happened to settle on better optima.

\paragraph{(RQ5) Testing population size.}
An analysis by Friedman et al.~\cite{Friedman2020} found that, using synthetic feature distributions, verification accuracy remained stable as population size increased whereas identification accuracy decreased.
Since we have access to real gaze data from thousands of people in the GazePro dataset, we wanted to test whether those results hold in practice.

This experiment (Experiment~\expBothBothFoldsFour{}) differs from the others in a few ways.
First, we used \num{4000}~users for the training set and the remaining \num{5202}~users for the testing set.
Second, we trained an ensemble of \num{4}~models using a \num{4}-fold split of the training set similar to~\cite{lohr_2022_ekyt}.
Third, we drew \num{100}~independent samples of $N$~users from the testing set without replacement, ensuring that each selected user had both recordings of interest so they could be both enrolled and verified.
We computed performance measures for each sample of $N$~users and then computed the P5 and P95 of each measure across all samples.
In Table~\ref{tab:highlighted-results}, for the sake of brevity, we reported the average of the P5 and P95 values.

\begin{figure}
    \centering
    \includegraphics[width=\linewidth]{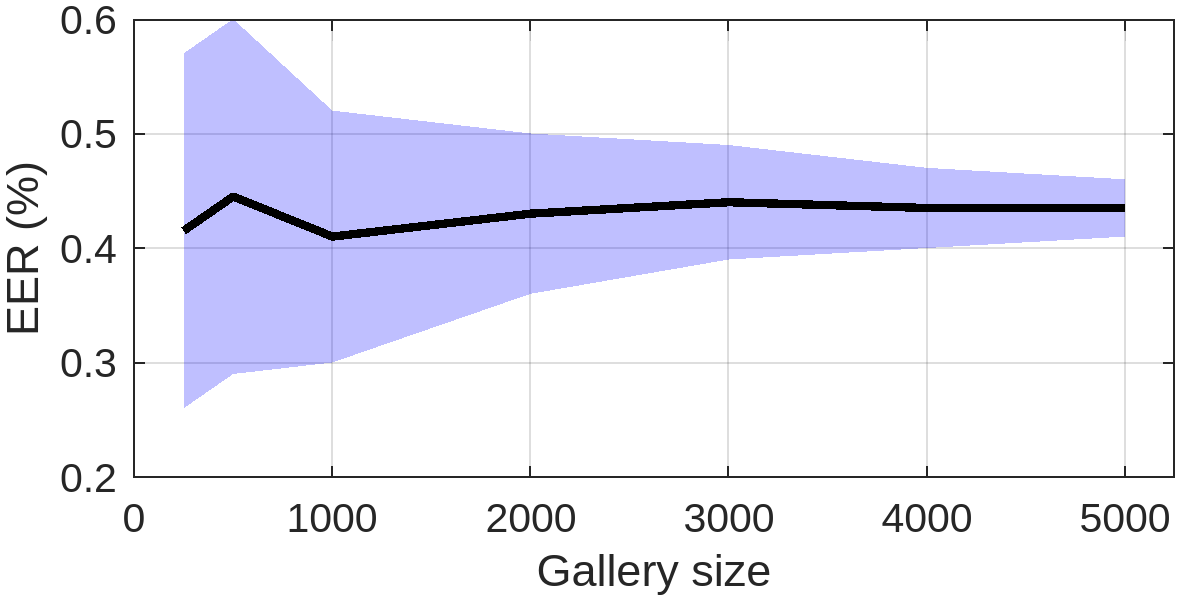} \\
    \includegraphics[width=\linewidth]{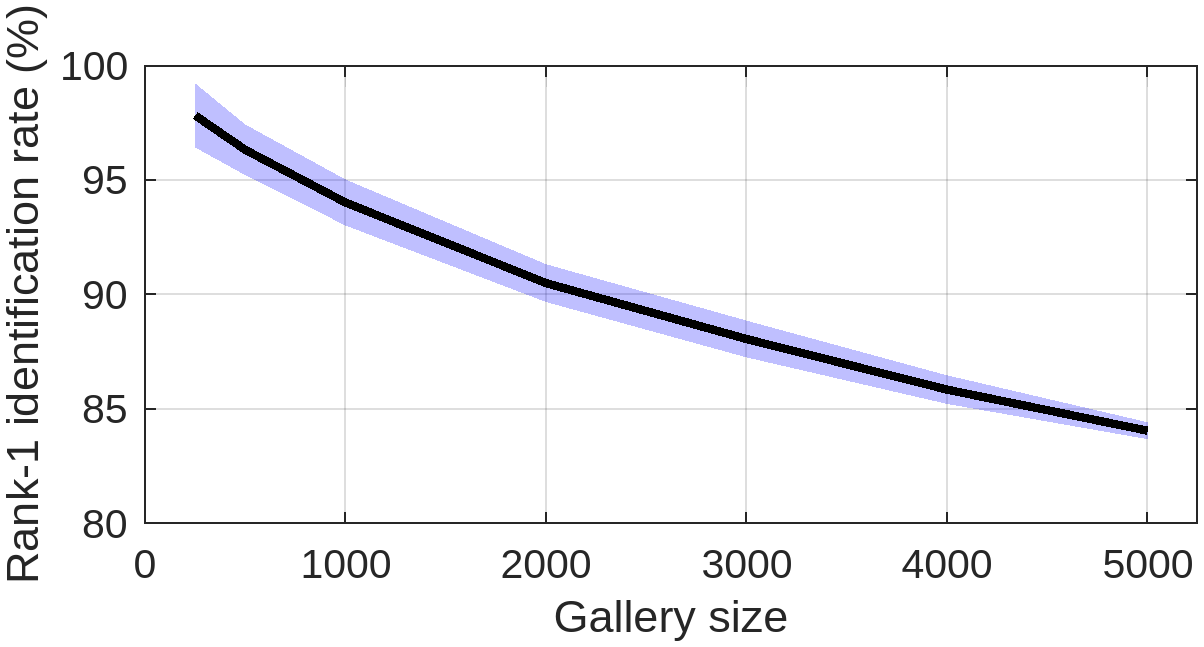}
    \caption{Performance measures vs gallery size from Experiment~\expBothBothFoldsFour{}.  The shaded region represents the \num{5}\textsuperscript{th} and \num{95}\textsuperscript{th} percentiles across \num{100}~random samples, and the black line is the average of the two percentiles.  Top: \gls{eer}.  Bottom: Rank-1 \gls{ir}.}
    \label{fig:perf-vs-users}
\end{figure}

\begin{figure}
    \centering
    \includegraphics[width=\linewidth]{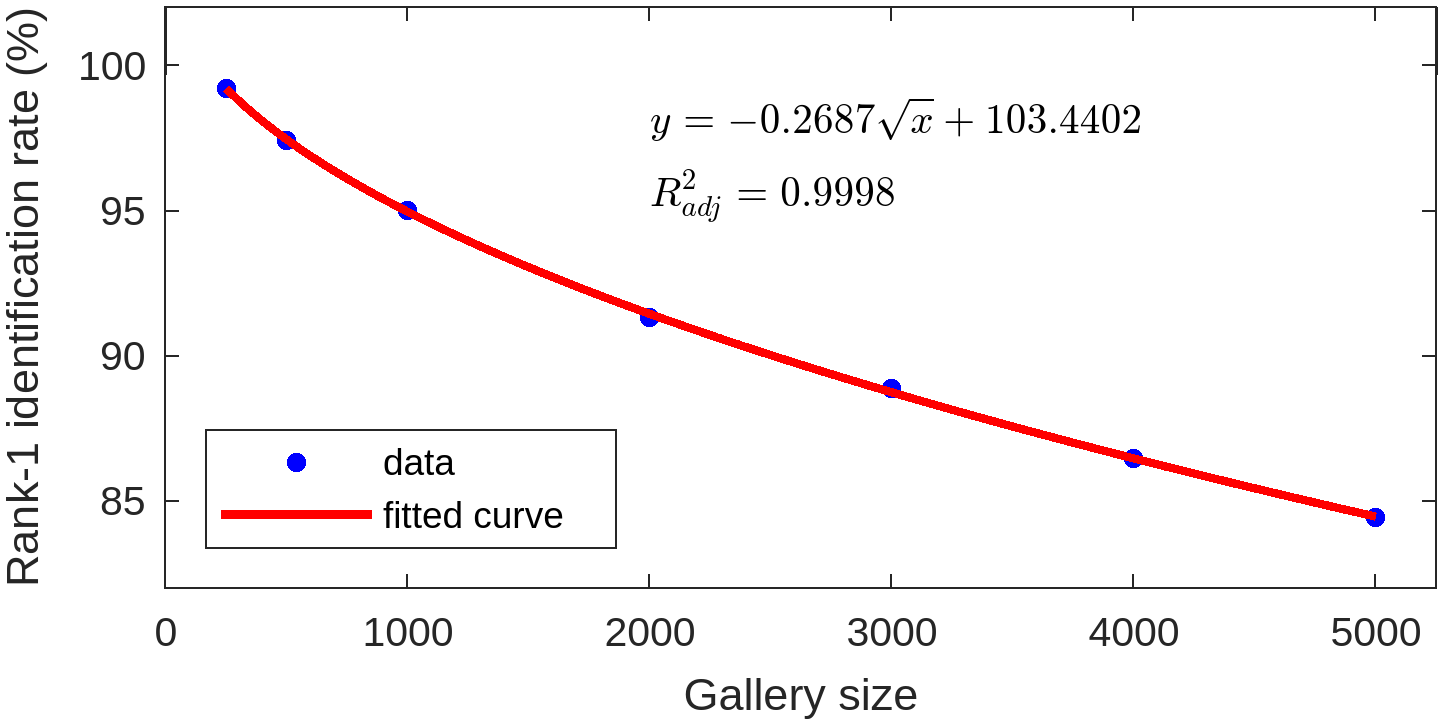}
    \caption{Identification rate (P95 across \num{100}~random samples) vs gallery size from Experiment~\expBothBothFoldsFour{}.  A curve of the form $a\sqrt{x}+b$ is fit to the observations.  Extrapolating from this curve, we roughly estimate that the identification rates might reach below-chance levels starting at gallery sizes of around \num{148000}~users.}
    \label{fig:fit-sqrt}
\end{figure}

As illustrated in Figure~\ref{fig:perf-vs-users}, we found that verification rates had a consistent central tendency regardless of population size, whereas the identification rates decreased as the population size increased.
These results offer practical validation of the theoretical analysis from Friedman et al.~\cite{Friedman2020}.
In Figure~\ref{fig:fit-sqrt}, we roughly estimated that the identification rate might decrease to below-chance levels when the population size increased to around \num{148000}~users.

\paragraph{(RQ6) Eye tracking signal quality.}
\Gls{et} signal quality over a large population of people varies for many reasons, including but not limited to eye color, eye shape, presence of corrective lenses, makeup, and device placement on a user's head~\cite{holmqvist2011comprehensive}.
Spatial accuracy, which is defined as the error between a user's actual gaze and the gaze vector estimated by a gaze estimation pipeline, is one of the most important \gls{et} signal quality characteristics.
Although a constant error bias in the gaze estimates would not affect measures of velocity, we consider that spatial accuracy is often dependent on gaze angle and might therefore affect features such as saccade curvature.

\begin{figure}[t]
    \centering
    \includegraphics[width=\linewidth]{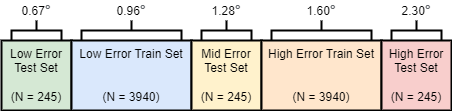}
    \caption{Partitions of a subset of users in GazePro based on their average spatial accuracy.  The degree value above a partition represents the median spatial accuracy across users in that partition.}
    \label{fig:accuracy-groups}
\end{figure}

The subset of \num{8615}~participants from the full dataset who were recorded for both random saccade tasks were partitioned based on their average spatial accuracy during the two tasks.
These partitions (illustrated in Figure~\ref{fig:accuracy-groups}) formed high error and low error train sets each with \num{3940}~participants, as well as high error, medium error, and low error test sets each with \num{245}~participants.

\Gls{eer} was highest on the high error test set for both Experiments~\expBothBothHigh{} and~\expBothBothLow{}, which may indicate that poor spatial accuracy reduces authentication performance.
Additionally, a model trained on data with high/low error performed worse when evaluated on data with low/high error.
However, we observed a different trend when using the visual axis alone instead of both axes (see the supplementary material).

\paragraph{(RQ7--8) Enrollment/verification duration.}
The prior experiments employed \num{20}~seconds of data for both enrollment and verification.
In this section, we employed presentation durations of \num{5}, \num{10}, and \num{15}~seconds to establish corresponding performance baselines.
To do this, instead of aggregating embeddings over the first four \num{5}-second chunks of a recording, we used the first one, two, or three chunks.
We used the model from Experiment~\expBothBothBig{} for this analysis.

Using the first \num{5}~seconds for enrollment and verification, we achieved \num{0.18}\% \gls{eer} and \num{41.9}\% $\text{FRR}_{0.002\%}$.
Using the first \num{10}~seconds for enrollment and verification, we achieved \num{0.08}\% \gls{eer} and \num{20.6}\% $\text{FRR}_{0.002\%}$.
Using the first \num{15}~seconds for enrollment and verification, we achieved \num{0.05}\% \gls{eer} and \num{8.8}\% $\text{FRR}_{0.002\%}$.

If we instead used a fixed enrollment duration of \num{30}~seconds, we observed a similar trend of authentication performance improving with increasing verification duration.
We observed that enrolling with \num{30}~seconds and verifying with $X$~seconds (up to $X=15$) yielded roughly similar results as enrolling and verifying with $X+5$~seconds.

\paragraph{(RQ9) Task-independence.}
The prior experiments used a random saccade (jumping dot) task for enrollment and verification.
How much, if at all, would performance change by using a smooth pursuit (gliding dot) task instead?
We used the same inputs and training/testing sets as Experiment~\expBothBoth{} for this analysis, but we trained an ensemble of \num{4}~models using a \num{4}-fold split of the training set similar to~\cite{lohr_2022_ekyt}.

With this configuration (Experiment~\expBothBothFolds{}), when enrolling and verifying with \num{20}~seconds from random saccade tasks, we achieved \num{0.10}\% \gls{eer} and \num{22.9}\% $\text{FRR}_{0.002\%}$.
If we instead enrolled and verified with \num{20}~seconds from smooth pursuit tasks, we achieved \num{0.10}\% \gls{eer} and \num{24.8}\% $\text{FRR}_{0.002\%}$.
These two sets of results were very similar.

\paragraph{(RQ10) Feature permanence.}
For this experiment, we employed the model from Experiment~\expBothBothBig{} to produce embeddings.
We then assessed the permanence (temporal persistence, reliability) of L2-normalized embedding features using the \gls{icc}, which has been applied to eye movement traits~\cite{Friedman2017,Friedman2020b,raju_2024_embeddings} and is also commonly used in other domains (e.g.,~\cite{Xue2021}).
We found that the median \gls{icc} of the \num{128}~embedding features was~\num{0.94}, with a minimum of~\num{0.90} and a maximum of~\num{0.98}.

\section{Discussion}
(RQ1) Binocular gaze data yielded a substantial improvement in verification accuracy over monocular gaze data (Experiments~\expLeftVisual{} and~\expBothVisual{}).
Having access to data from both eyes provides estimates of eye dominance and vergence angle, and the model may be able to utilize this information to increase performance.

(RQ2) We also found that including both the optical and visual axes as inputs to the model yields a substantial improvement in verification accuracy over using the visual axis alone (Experiments~\expBothVisual{} and~\expBothBoth{}).
The difference between the visual and optical axes would essentially provide an estimate of a user's calibration parameters, and this difference alone achieves good verification accuracy (Experiment~\expBothDiff{}).
These results suggest that the angle kappa between the optical and visual axes might be a highly distinctive trait.
In a scenario such as continuous authentication where a calibration period could not precede every verification attempt, it may be feasible to employ a cached set of calibration parameters from the enrolled user to estimate the visual axis.

(RQ3) By training for up to \num{10}-times longer and with a \num{4}-times larger minibatch size, we achieved our best-case result of \num{2.4}\%~$\text{FRR}_{0.002\%}$ (Experiment~\expBothBothBigFolds{}).
This result is especially noteworthy since, using just \num{20}~seconds of \num{72}~Hz gaze data, we were able to meet the FIDO requirements~\cite{FIDO2024} of \num{3}\% \gls{frr} at a \num{1}-in-\num{50000} \gls{far} with a maximum verification duration of \num{30}~seconds.
It is important to note that these results are for a short test-retest interval, and we would expect performance to degrade over longer intervals as past studies have shown.
Additional user studies are needed to understand whether \num{20}~seconds would be tolerable for users of a device unlock feature in an \gls{xr} form factor.

(RQ4) We observed an increase in verification rates as the training population size increased from \num{250}~to \num{6000}~users (Experiments~\expBothBothQuarter{}--\expBothBothSix{}), with the largest performance improvement occurring between \num{2000}~and \num{3000}~users.
These results suggest that, if there is a point at which performance stops improving when more users are included in the training set, such a point would be beyond \num{6000}~users, and training on an even larger population may yield even better verification rates.

(RQ5) We empirically confirmed in practice that, as the test population size increases, verification rates remain stable whereas identification rates decrease (Experiment~\expBothBothFoldsFour{}).
Eye movement traits are behavioral rather than physical, so they are expected to lack the distinctiveness necessary for large-scale identification.

(RQ6) Regarding the impact of spatial accuracy on authentication performance, we found that in some cases poor spatial accuracy reduced authentication performance (Experiments~\expBothBothHigh{}--\expBothBothLow{}), in contrast to a study by Raju et al.~\cite{raju_2024_embeddings}.
Our results also suggest that, for the best performance, the spatial accuracy of the training dataset should be representative of the level of spatial accuracy present in the testing dataset.
Further research into the relationship between authentication performance and spatial accuracy is necessary.

(RQ7--8) We observed that, using the model from Experiment~\expBothBothBig{}, $\text{FRR}_{0.002\%}$ is approximately halved for every additional \num{5}~seconds used for enrollment and verification, starting at \num{41.9}\% with \num{5}~seconds and improving to \num{4.1}\% with \num{20}~seconds.
Additionally, we observed that increasing the enrollment duration improves performance for shorter verification durations, potentially reducing the time needed for verification attempts at the cost of a longer initial enrollment procedure.
It is important to note, though, that user calibration parameters have still been determined from a full-length calibration process.

(RQ9) We found that, when using both the optical and visual axes as inputs to the model, enrolling and verifying with a smooth pursuit task results in similar performance as using a random saccade task (Experiment~\expBothBothFolds{}).
This is evidence that gaze-based verification remains to be task-independent even for larger population sizes.

(RQ10) All 128~learned embedding features of the model from Experiment~\expBothBothBig{} have \glspl{icc} at or above \num{0.9}, indicating that they are highly reliable according to all but the strictest of the commonly used thresholds~\cite{Xue2021}.
This supports the idea from Friedman et al.~\cite{Friedman2017} that higher \glspl{icc} are correlated with higher authentication accuracies.

Further discussion of the limitations of this work can be found in the supplementary material.

\section{Conclusion}
In this paper, we investigated authentication performance with a large dataset of gaze recordings collected in \gls{vr}.
Our most important findings indicate that the amount of training data is critical to achieve the authentication performance that is expected from modern wearable devices.
Given a very large dataset, which in our case was nearly \num{10000}~people, it is important to have a longer training time and increased minibatch size.
We have highlighted that the best authentication performance can be achieved if binocular estimates of both the optical and visual axes are provided.
Crucially, and in support of prior theoretical findings, we found that gaze data alone cannot be employed for (re-)identification purposes if the user pool is sufficiently large.
We also discovered the impact of spatial accuracy, an important descriptor of \gls{et} signal quality, such that higher signal quality resulted in better authentication performance.
In addition, we reinforced the importance of data length on resulting accuracy, specifically that longer enrollment/verification durations improved authentication accuracy.
We also found that gaze-based authentication in our study appeared to be task-independent.
In summary, gaze authentication can be as accurate as required by the FIDO standard when driven by a state-of-the-art deep learning architecture and a sufficiently large training dataset.

\section*{Acknowledgments}
We thank Meta for providing us with access to the GazePro dataset and the compute resources necessary to perform this important work.

{\small
\bibliographystyle{ieee}
\bibliography{ms}
}

\end{document}


\title{Supplementary Material for ``Establishing a Baseline for Gaze-driven Authentication Performance in VR: A Breadth-First Investigation on a Very Large Dataset''}

\author{Dillon Lohr$^{1,2}$, Michael J. Proulx$^{2}$, Oleg Komogortsev$^{1,2}$\\
$^{1}$Texas State University, San Marcos, TX, USA\\
$^{2}$Meta Reality Labs Research, Redmond, WA, USA\\
{\tt\small djl70@txstate.edu, michaelproulx@meta.com, ok@txstate.edu}}

\maketitle
\thispagestyle{empty}

\section{Purpose of this document}
This supplementary material contains additional information about the dataset, methodology, results, and limitations of our study that would not fit in the main manuscript.

\section{Methodology}

\subsection{Dataset}

Of the \num{9202}~participants in GazePro, \num{3673}~self-reported as male, \num{5376}~as female, and \num{153}~as neither male nor female.
The participants' ages ranged from \num{13}~to \num{88}~years (median=\num{33}, IQR=\num{18}).
The self-reported race/ethnicity of participants is given in Table~\ref{tab:race-ethnicity}.
All participants had normal or corrected-to-normal vision, with \num{3223}~wearing glasses or contact lenses and \num{5979}~wearing no corrective lenses.
There were \num{2032}~participants that wore some form of makeup around the eyes, including eyelash extensions, heavy eyeshadow, and/or mascara.
In terms of eye color, \num{5614}~participants had brown eyes, \num{1759}~blue, \num{1113}~hazel, \num{701}~green, \num{13}~amber, and \num{2}~violet.

The data in GazePro were collected in a specialized facility equipped for such recordings.
Subjects did not use a chin rest.
Tasks included random saccade tasks, smooth pursuit tasks, and vestibulo-ocular reflex tasks, each with a single focal target designed to elicit a high degree of focus.
Total duration of all tasks was approximately \num{20}~minutes.
Some tasks were similar to those found in the study by Lohr et al.~\cite{lohr_2023_gazebasevr}.
Luminosity of the background was changed similarly to the study by Aziz et al.~\cite{aziz_2024_evaluation}.
During some tasks, users were asked to make facial expressions such as winking, blinking, squinting, laughing, being angry, and being sad.
At certain times, subjects were asked to adjust---and thus slightly shift---the position of the \gls{vr} device on their heads.
Throughout the procedure, an experiment facilitator was nearby to help achieve high compliance with the task description.

\begin{figure*}
    \centering
    \includegraphics[width=\linewidth]{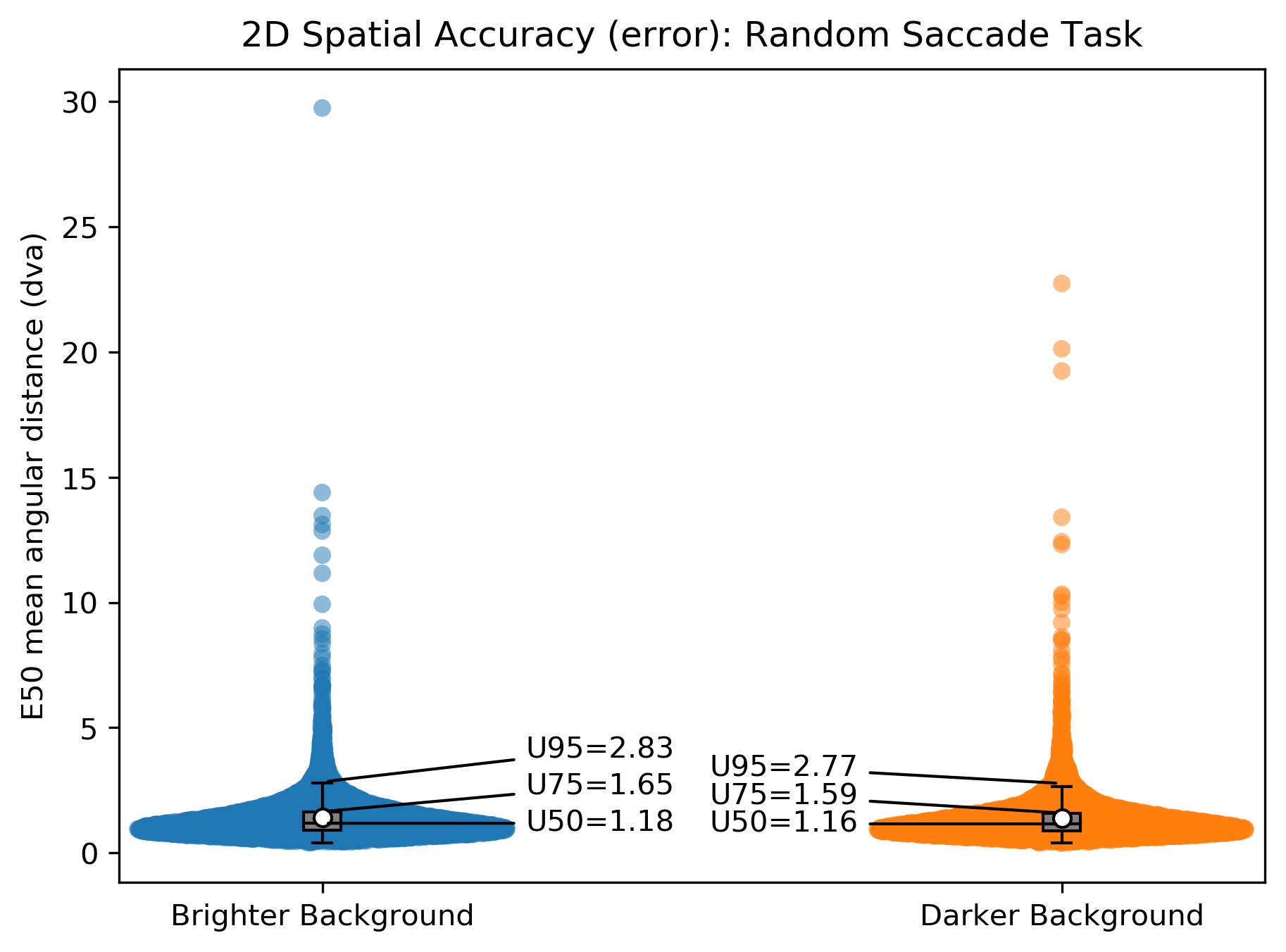}
    \caption{Distributions of median (E50) spatial accuracy across the N=\num{8615}~participants who were recorded for both random saccade tasks.  There is one blue and orange dot per participant.  The 50th, 75th, and 95th user percentiles are labeled as U50, U75, and U95, respectively.  The white dot represents the mean of the distribution.  The boxplot covers U25 to U75 with a line indicating U50, and the whiskers extend to $\text{U25} - 1.5 \times \text{IQR}$ and $\text{U75} + 1.5 \times \text{IQR}$.}
    \label{fig:e50-acc}
\end{figure*}

\begin{figure*}
    \centering
    \includegraphics[width=\linewidth]{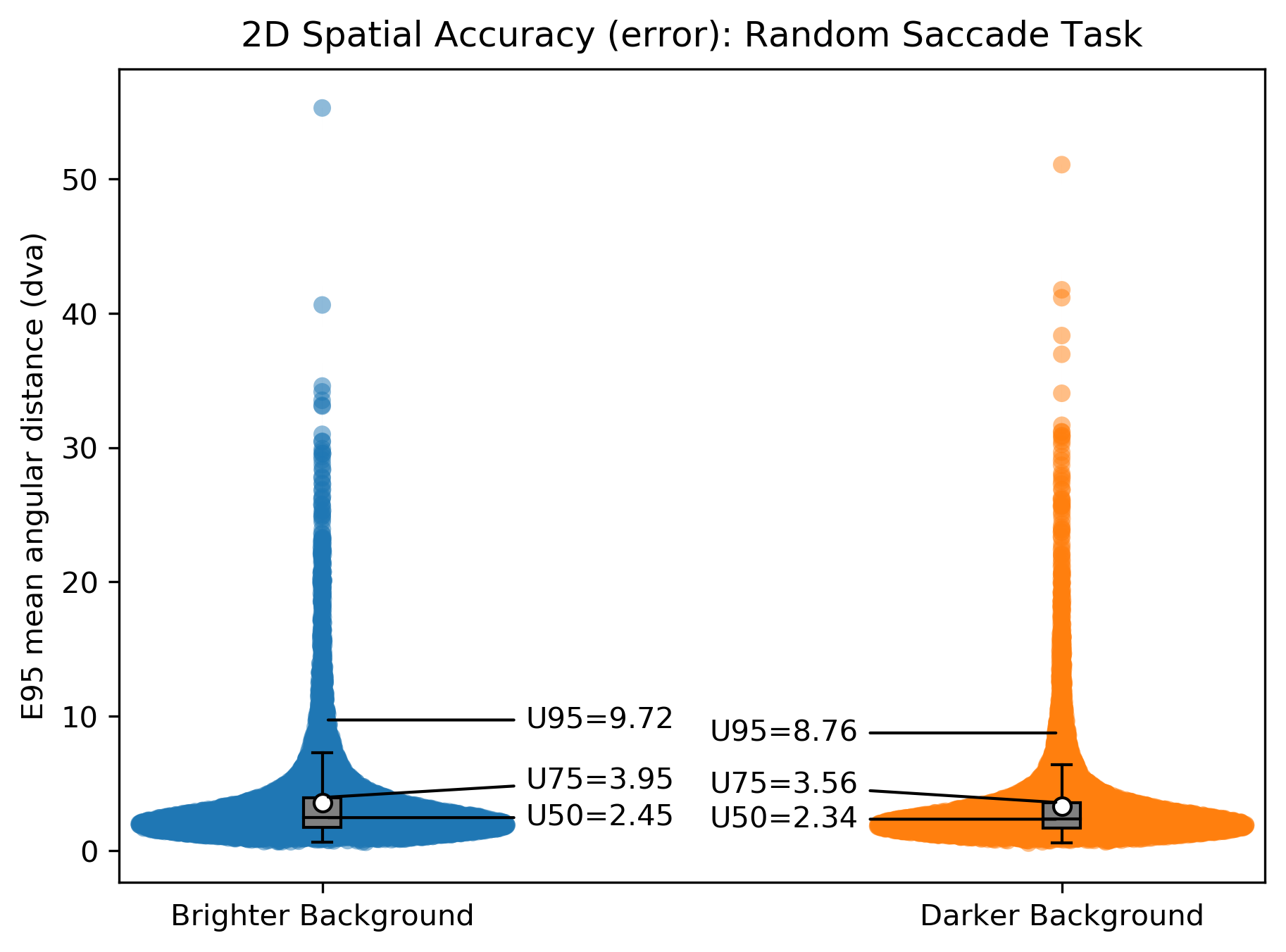}
    \caption{Distributions of 95th percentile (E95) spatial accuracy across the N=\num{8615}~participants who were recorded for both random saccade tasks.  There is one blue and orange dot per participant.  The 50th, 75th, and 95th user percentiles are labeled as U50, U75, and U95, respectively.  The white dot represents the mean of the distribution.  The boxplot covers U25 to U75 with a line indicating U50, and the whiskers extend to $\text{U25} - 1.5 \times \text{IQR}$ and $\text{U75} + 1.5 \times \text{IQR}$.}
    \label{fig:e95-acc}
\end{figure*}

GazePro was collected on a \gls{vr} headset with \gls{et} signal quality and other general characteristics equivalent to a Meta Quest Pro~\cite{aziz_2024_evaluation,Wei2023}.
Figures~\ref{fig:e50-acc} and~\ref{fig:e95-acc} visualize distributions of spatial accuracy across the subset of \num{8615}~participants who were recorded for both random saccade tasks.

\subsection{Network architecture}
The \gls{ekyt} architecture consists of a single DenseNet-style~\cite{huang2018densely} block of \num{8}~convolution layers, where the feature maps output from one layer are concatenated with the outputs of all previous layers before being passed into the next layer.
This DenseNet block is followed by global average pooling and a single fully-connected layer.
All convolution layers but the first are preceded by \gls{bn}~\cite{ioffe2015batchnorm} and the \gls{relu} activation function~\cite{nair2010relu}.
\Gls{bn} and \gls{relu} are also applied to the final set of concatenated feature maps prior to global average pooling.

\subsection{Model training}
Each \gls{et} recording is pre-processed following the methodology from~\cite{lohr_2022_ekyt}.
First, a recording is partitioned into non-overlapping \num{5}-second chunks.
Next, velocity is computed with a Savitzky-Golay~\cite{savitzky1964sgolay} differentiation filter with order 2 and window size 7.
Then, velocities are clamped to $\pm 1000^{\circ}/\text{s}$.
Finally, clamped velocities are standardized to have zero mean and unit variance using the formula $(x - \mu) / \sigma$, where $\mu$ and $\sigma$ are the mean and \gls{sd}, respectively, of the training set.

The model is trained using \gls{ms} loss~\cite{Wang2019}.
\Gls{ms} loss is formulated as
\begin{equation}
\begin{split}
    L_{MS} =\, &\frac{1}{m} \sum_{i = 1}^{m} \left( \frac{1}{\alpha} \log \left( 1 + \sum_{k \in P_i} \exp \left( -\alpha \left( S_{ik} - \lambda \right) \right) \right) \right. \\
    &+ \left. \frac{1}{\beta} \log \left( 1 + \sum_{k \in N_i} \exp \left( \beta \left( S_{ik} - \lambda \right) \right) \right) \right),
\end{split}
\end{equation}
where $m = 256$ is the size of each minibatch; $\alpha = 2.0$, $\beta = 50.0$, and $\lambda = 0.5$ are hyperparameters for \gls{ms} loss; $P_i$ and $N_i$ are the sets of indices of the mined positive and negative pairs for each anchor sample $\mathbf{x}_i$; and $S_{ik}$ is the cosine similarity between the pair of samples $\{ \mathbf{x}_i,\, \mathbf{x}_k \}$.
Note that this formulation of \gls{ms} loss implicitly includes an online pair miner with an additional hyperparameter $\varepsilon = 0.1$.
More details about \gls{ms} loss can be found in~\cite{Wang2019}.

Minibatches are constructed by randomly selecting \num{16}~unique users from the training set and \num{16}~unique samples for each selected user.
This yields $16 \times 16 = 256$ samples per minibatch.
Each training ``epoch'' iterates over as many minibatches as needed until we have selected a total number of samples equivalent to the number of unique samples in the training set.
Note that not every sample from the training set may be included in any given epoch.
Additionally, in some experiments, we construct minibatches using $32 \times 32 = 1024$ samples.

The model is trained for \num{100}~epochs using the Adam~\cite{kingma2014adam} optimizer with a one-cycle cosine annealing learning rate scheduler~\cite{smith2018superconvergence}.
With this learning rate schedule, the learning rate begins at $10^{-4}$, gradually increases to a maximum of $10^{-2}$ over the first \num{30}~epochs, then gradually decreases to a minimum of $10^{-7}$ over the subsequent \num{70}~epochs.

\subsection{Model evaluation}

\subsubsection{Verification}
\label{sec:verification}
At evaluation time, we select one set of \gls{et} recordings (at most one per person) to be used for enrollment and another set of \gls{et} recordings (at most one per person) to be used for verification.
Each \gls{et} recording is pre-processed in the same way as the training data: partition into non-overlapping \num{5}-second chunks, compute velocity with a Savitzky-Golay~\cite{savitzky1964sgolay} differentiation filter, clamp velocities to $\pm 1000^{\circ}/\text{s}$, and scale using the mean and \gls{sd} that were computed across the training set.
We note that the gaze estimation pipeline employed for GazePro provides valid gaze estimates for all data points even when the eye is blinking, so we did not include the minimum validity criteria from~\cite{lohr_2022_ekyt} as it was not necessary in this case.
The first $n$ of these processed chunks are independently fed into our model to produce $n$ embeddings.
Those $n$ embeddings are then averaged to produce a single centroid embedding.
After repeating this process for each recording, we have one set of centroid embeddings for enrollment and another set of centroid embeddings for verification.

We then simulate all possible verification attempts by comparing each centroid embedding from the verification set against each centroid embedding from the enrollment set.
Embeddings are compared using cosine similarity.
A verification attempt is \textit{genuine} if the two centroid embeddings originate from the same person or an \textit{impostor} if they originate from two different people.
The resulting similarity scores and labels are used to produce a \gls{roc} curve.

\begin{figure*}
    \centering
    \includegraphics[width=0.49\linewidth]{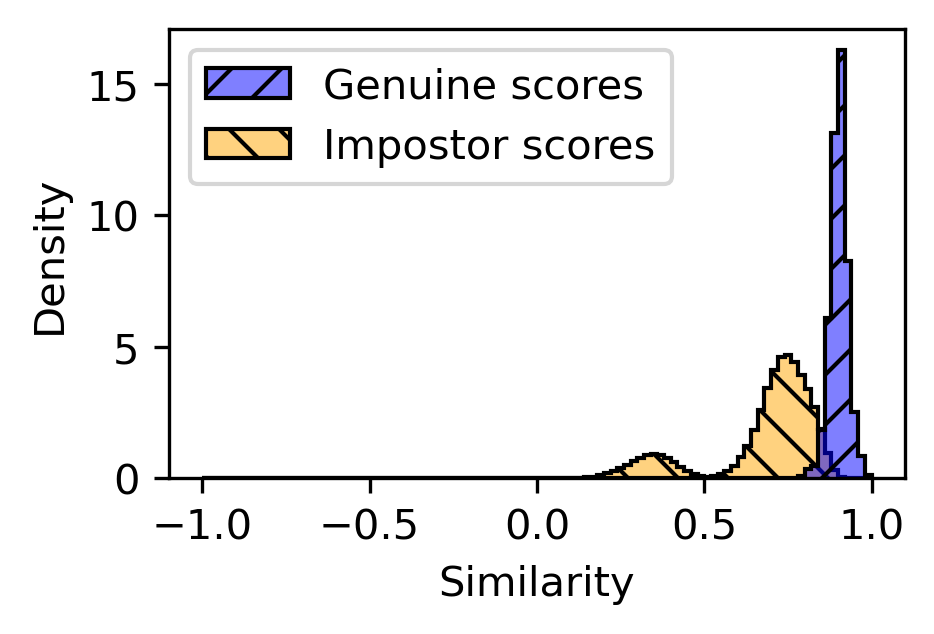}
    \includegraphics[width=0.49\linewidth]{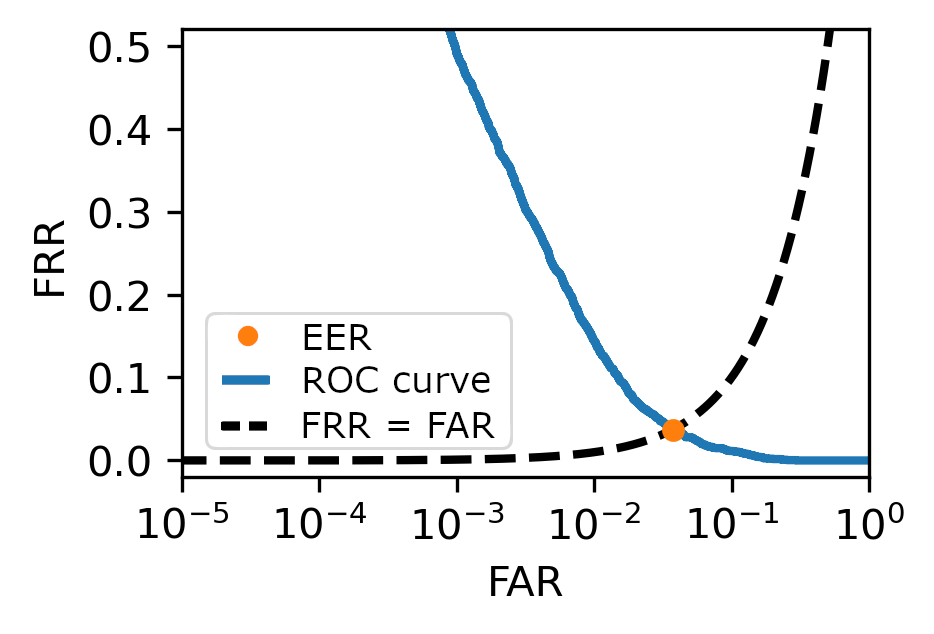}
    \caption{Experiment~\expBothVisual{} using 20~seconds for enrollment and verification.  Left: histograms of the genuine and impostor similarity score distributions.  Right: the ROC curve, with an orange dot indicating the EER and a dashed black line where $\text{FRR} = \text{FAR}$.}
    \label{fig:sim-roc-exp58}
\end{figure*}

\begin{figure*}
    \centering
    \includegraphics[width=0.49\linewidth]{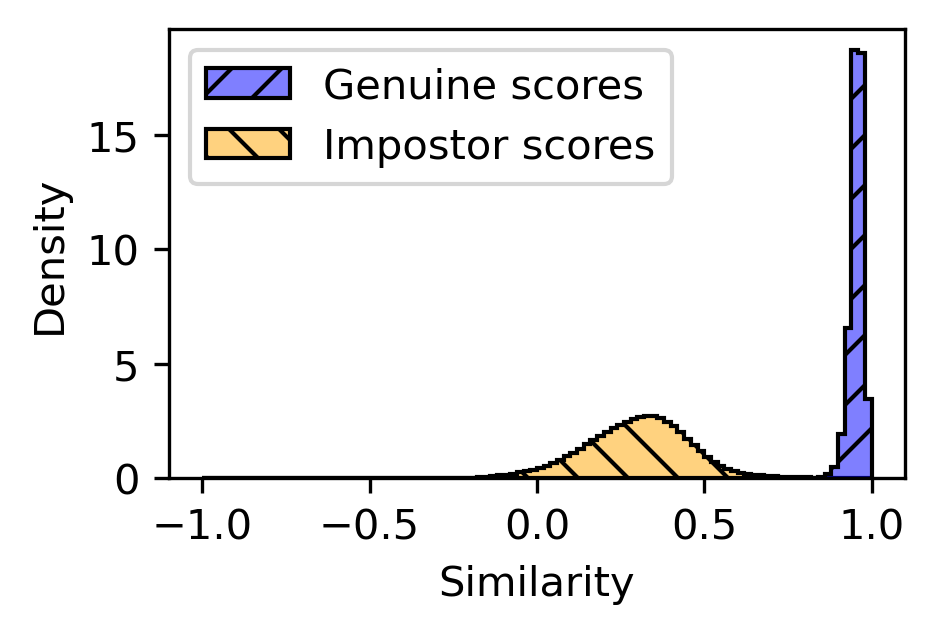}
    \includegraphics[width=0.49\linewidth]{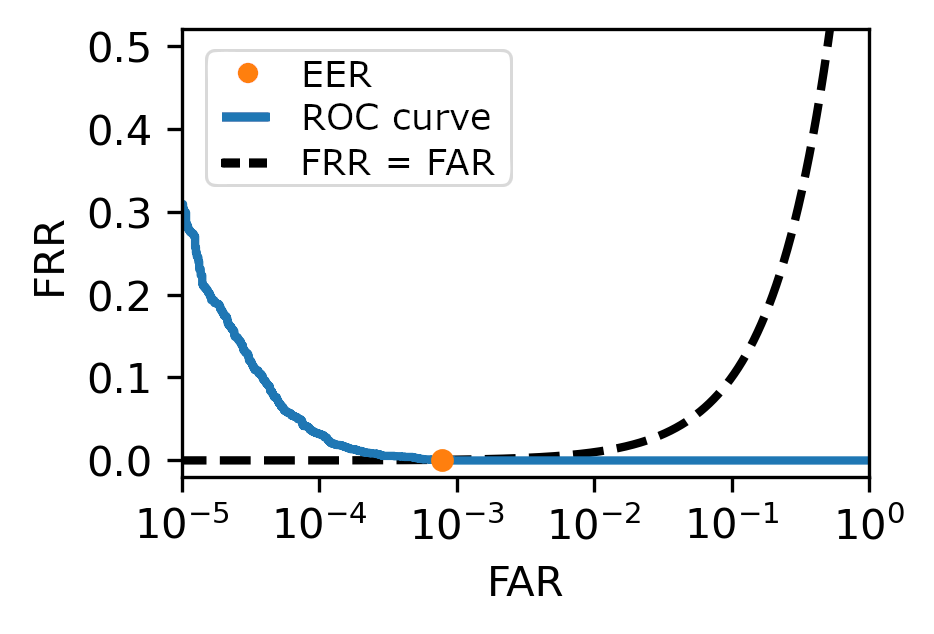}
    \caption{Experiment~\expBothBoth{} using 20~seconds for enrollment and verification.  Left: histograms of the genuine and impostor similarity score distributions.  Right: the ROC curve, with an orange dot indicating the EER and a dashed black line where $\text{FRR} = \text{FAR}$.}
    \label{fig:sim-roc-exp69}
\end{figure*}

\begin{figure*}
    \centering
    \includegraphics[width=0.49\linewidth]{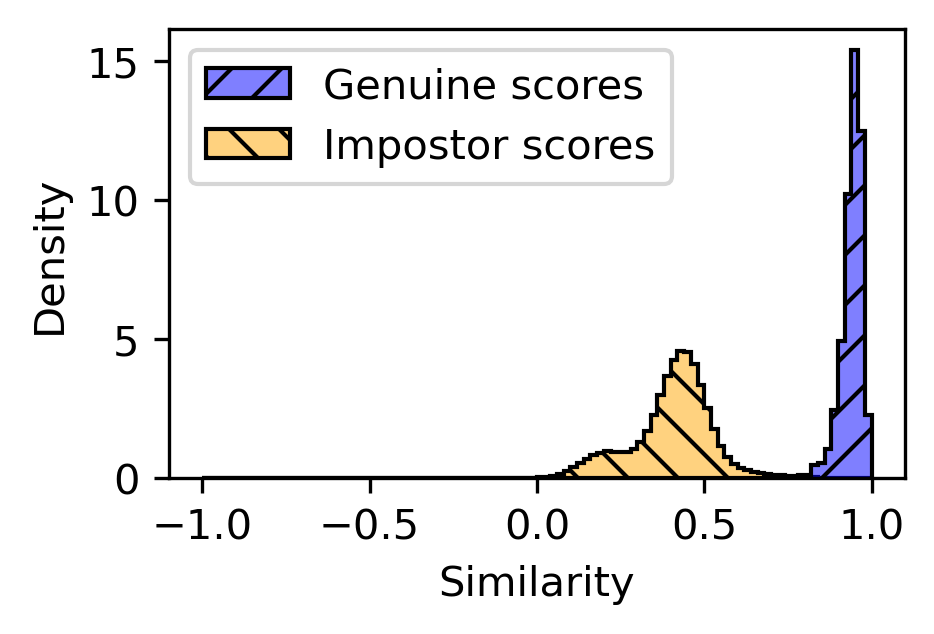}
    \includegraphics[width=0.49\linewidth]{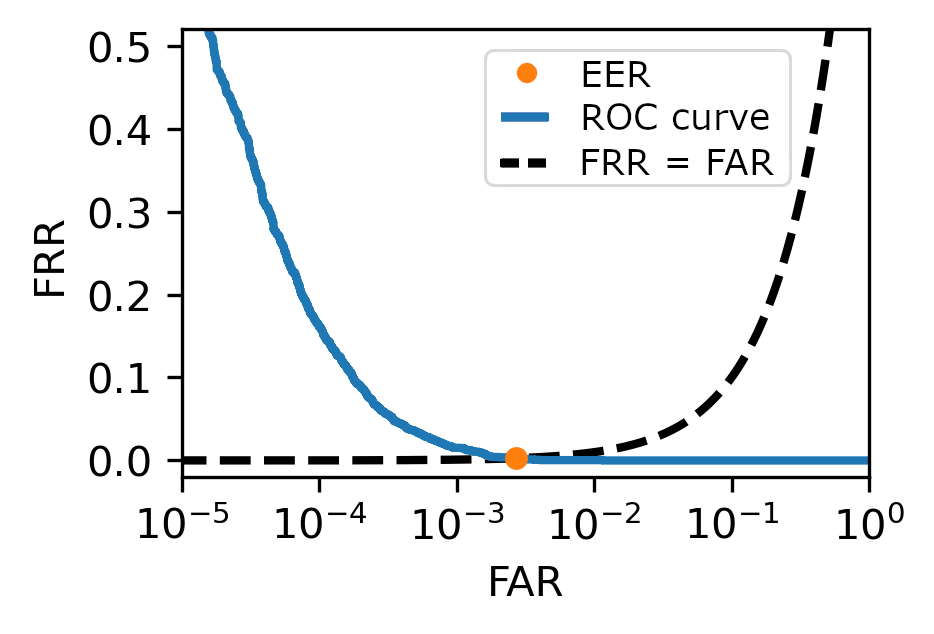}
    \caption{Experiment~\expBothDiff{} using 20~seconds for enrollment and verification.  Left: histograms of the genuine and impostor similarity score distributions.  Right: the ROC curve, with an orange dot indicating the EER and a dashed black line where $\text{FRR} = \text{FAR}$.}
    \label{fig:sim-roc-exp80}
\end{figure*}

\begin{figure*}
    \centering
    \includegraphics[width=0.49\linewidth]{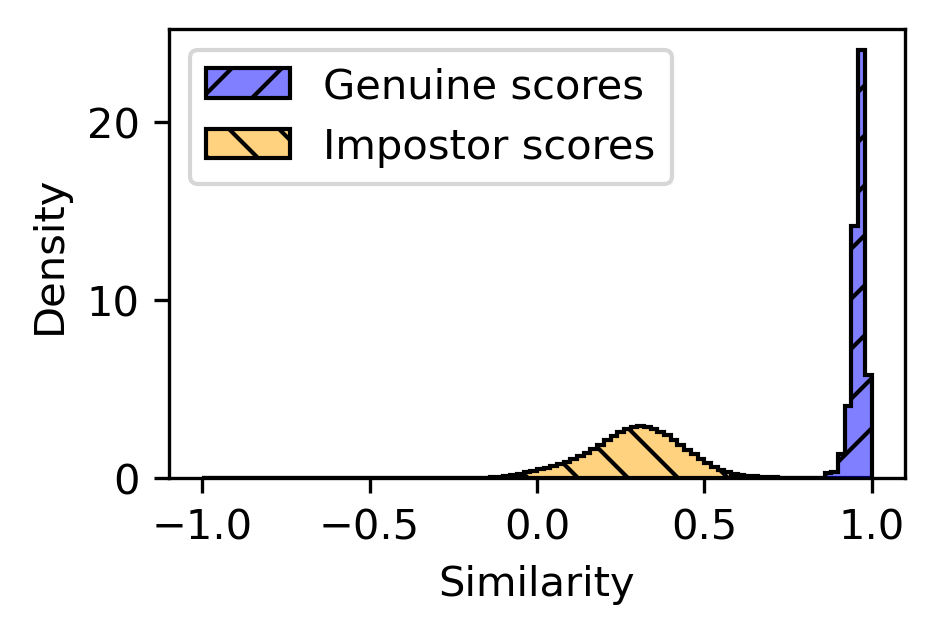}
    \includegraphics[width=0.49\linewidth]{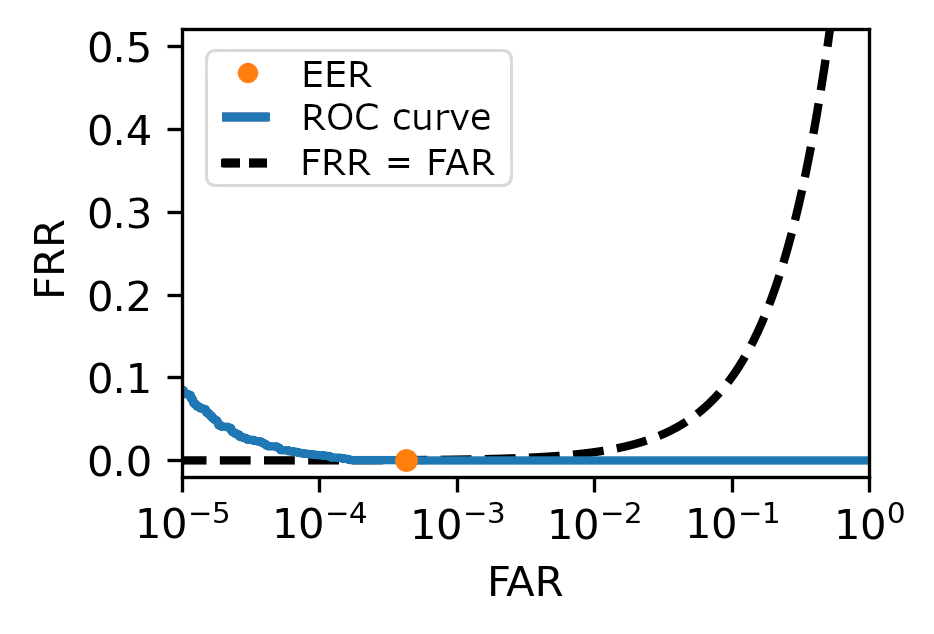}
    \caption{Experiment~\expBothBothBig{} using 20~seconds for enrollment and verification.  Left: histograms of the genuine and impostor similarity score distributions.  Right: the ROC curve, with an orange dot indicating the EER and a dashed black line where $\text{FRR} = \text{FAR}$.}
    \label{fig:sim-roc-exp54}
\end{figure*}

\begin{figure*}
    \centering
    \includegraphics[width=0.49\linewidth]{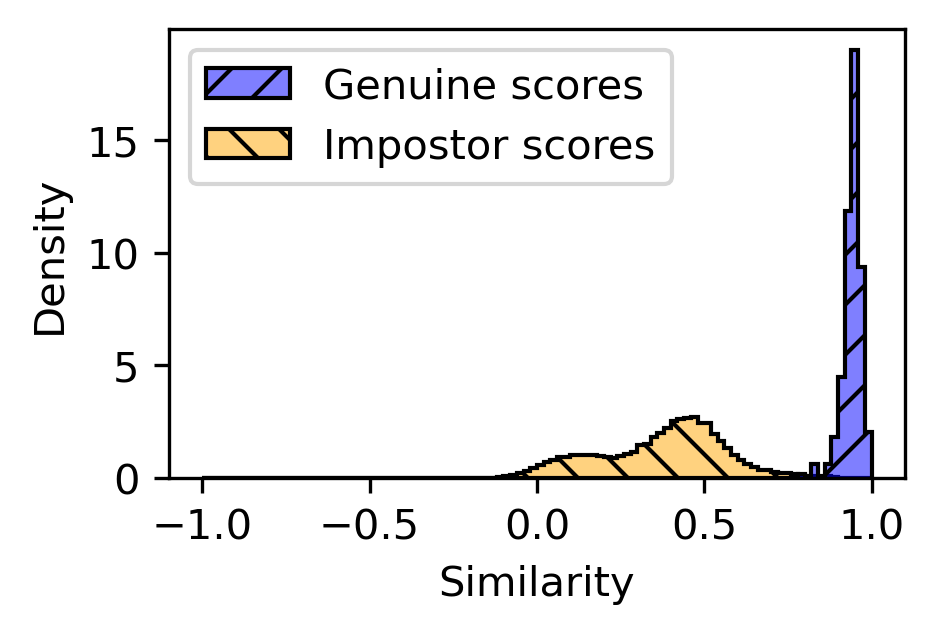}
    \includegraphics[width=0.49\linewidth]{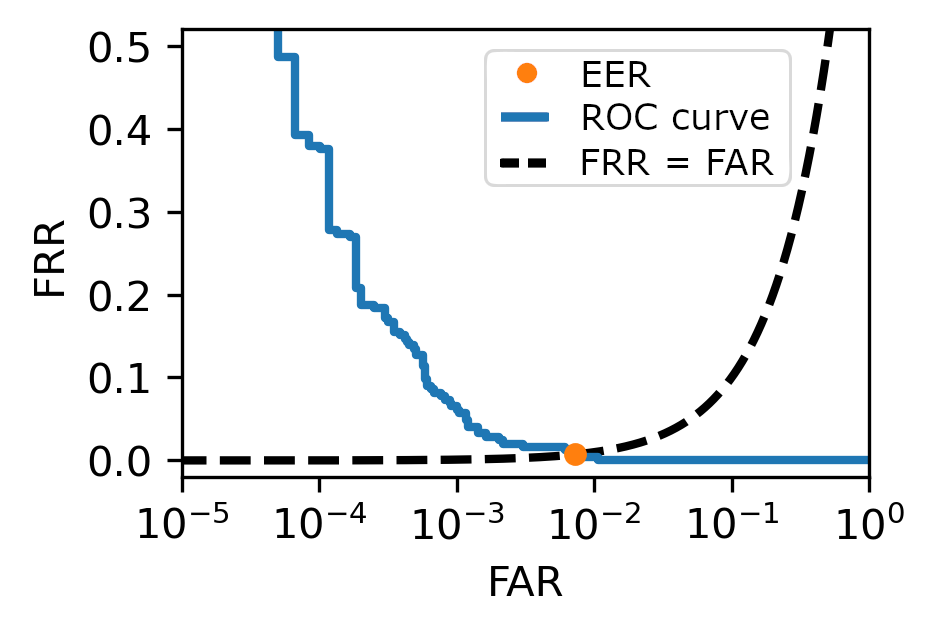} \\
    \includegraphics[width=0.49\linewidth]{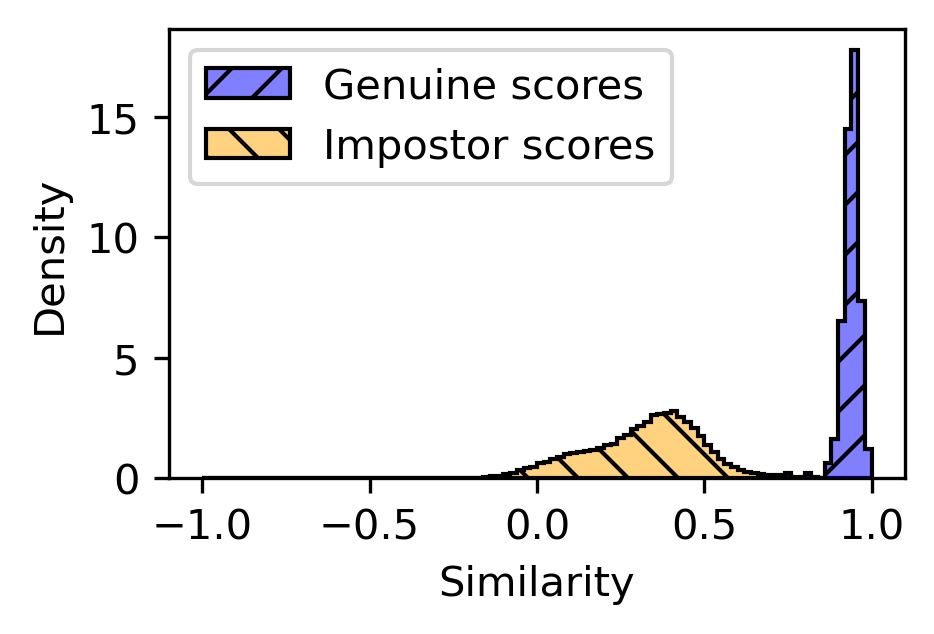}
    \includegraphics[width=0.49\linewidth]{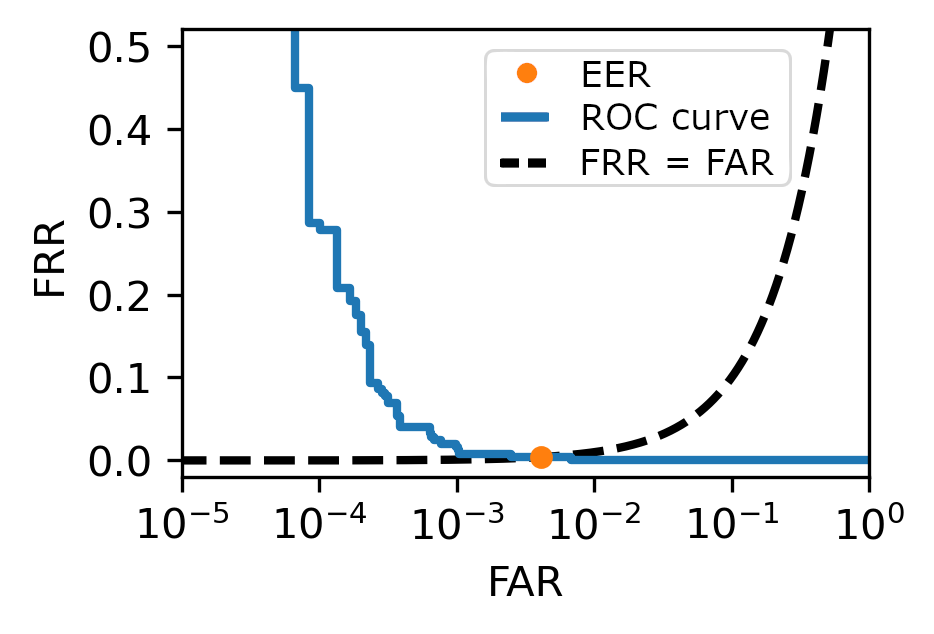} \\
    \includegraphics[width=0.49\linewidth]{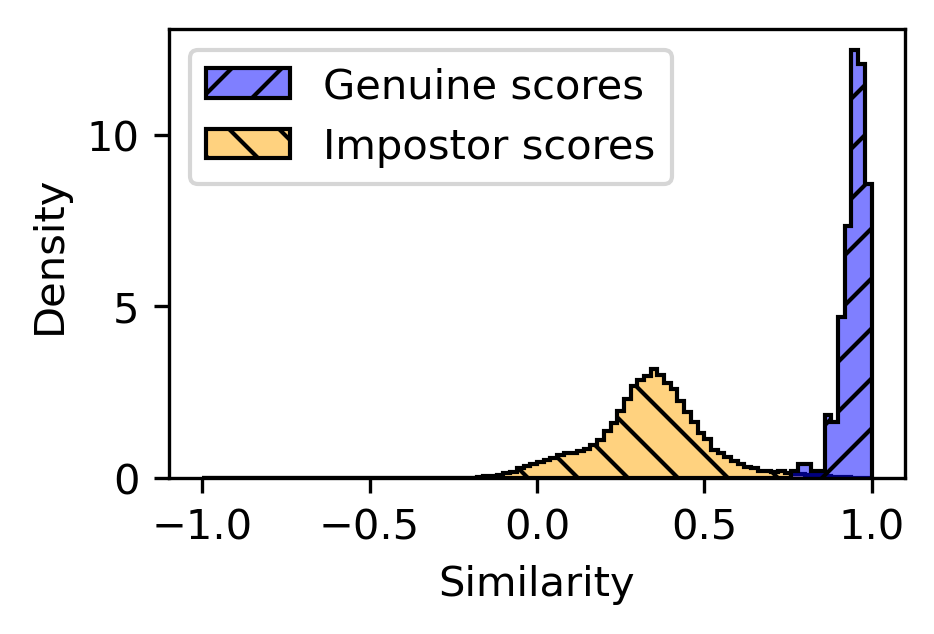}
    \includegraphics[width=0.49\linewidth]{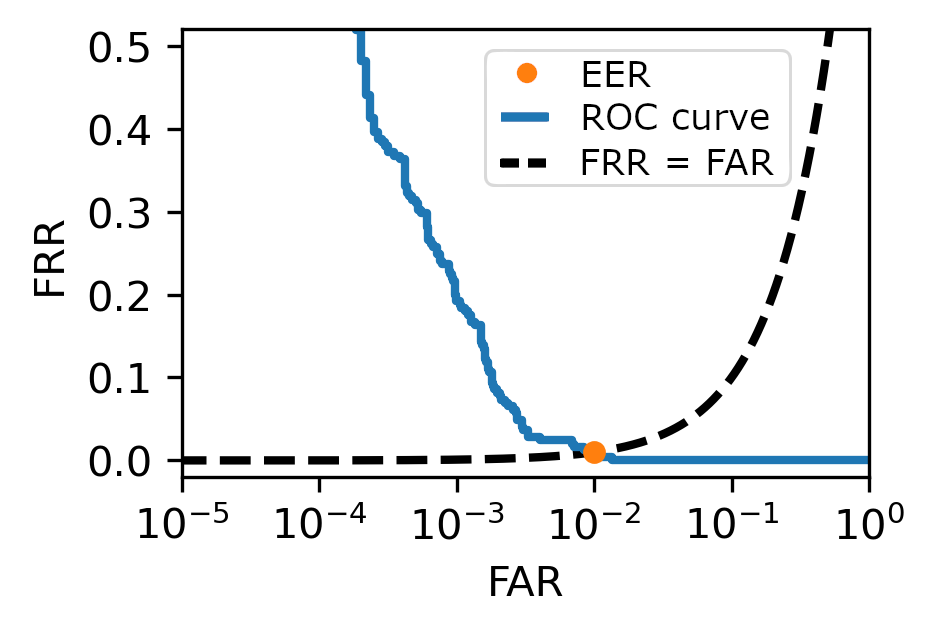} \\
    \caption{Experiment~\expBothBothHigh{} using 20~seconds for enrollment and verification.  Left: histograms of the genuine and impostor similarity score distributions.  Right: the ROC curve, with an orange dot indicating the EER and a dashed black line where $\text{FRR} = \text{FAR}$.  The top figures are with the high error test set, the center figures are with the mid error test set, and the bottom figures are with the low error test set.}
    \label{fig:sim-roc-exp70}
\end{figure*}

\begin{figure*}
    \centering
    \includegraphics[width=0.49\linewidth]{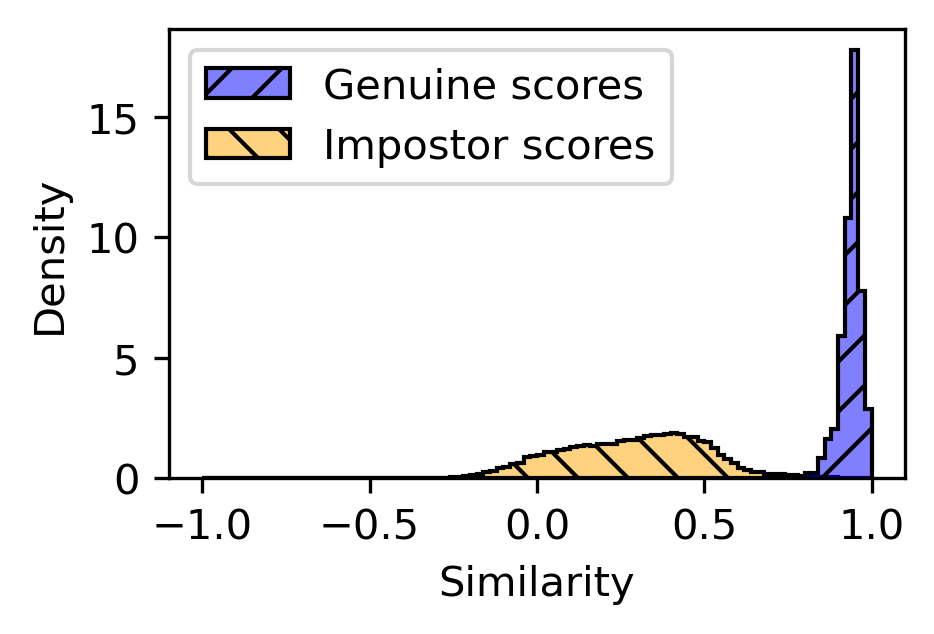}
    \includegraphics[width=0.49\linewidth]{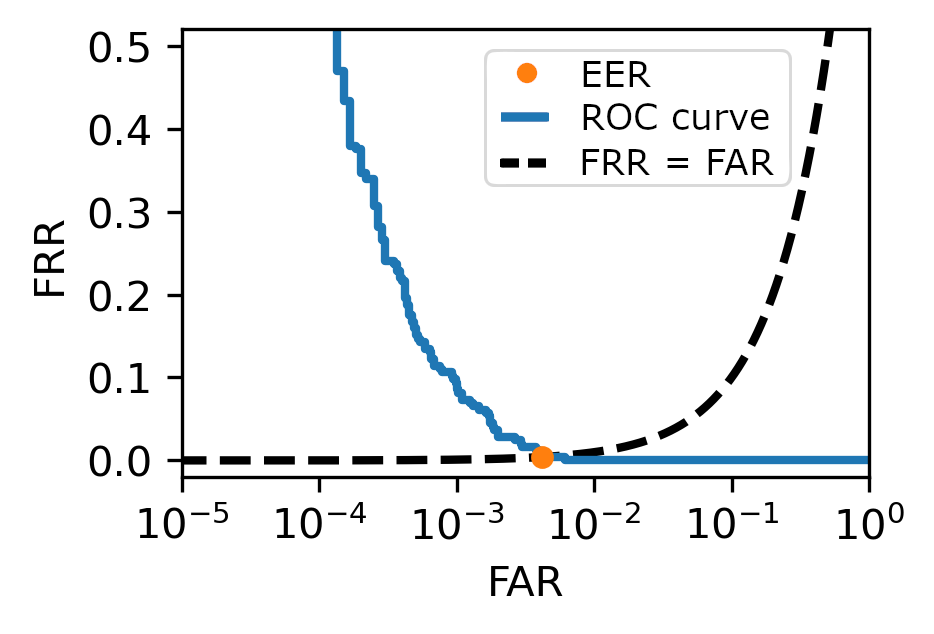} \\
    \includegraphics[width=0.49\linewidth]{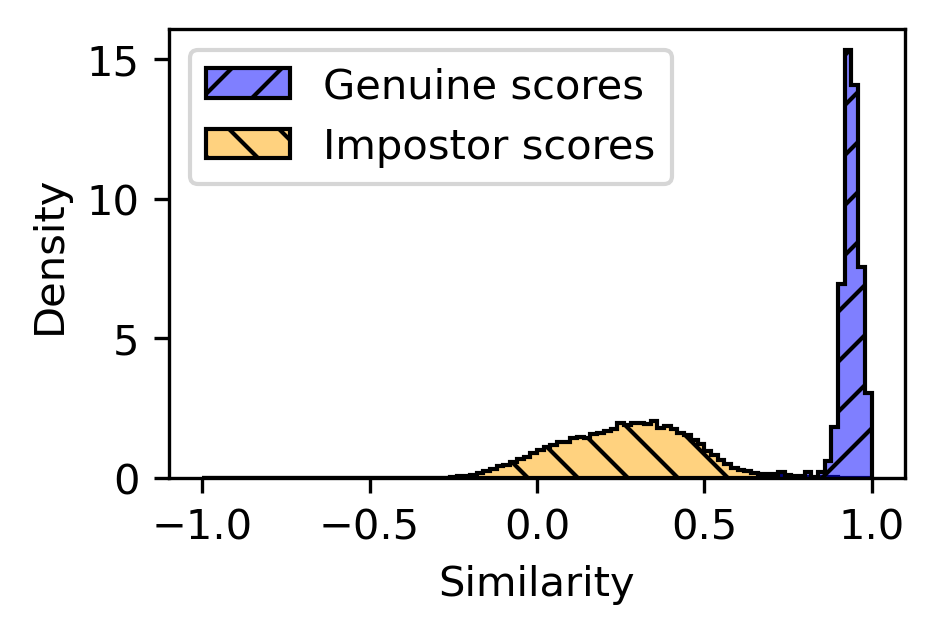}
    \includegraphics[width=0.49\linewidth]{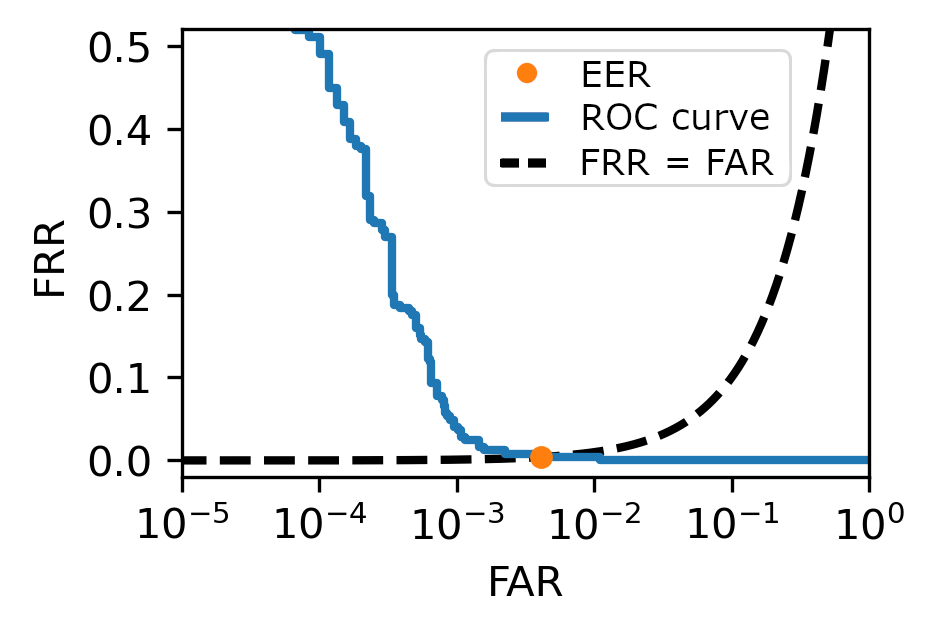} \\
    \includegraphics[width=0.49\linewidth]{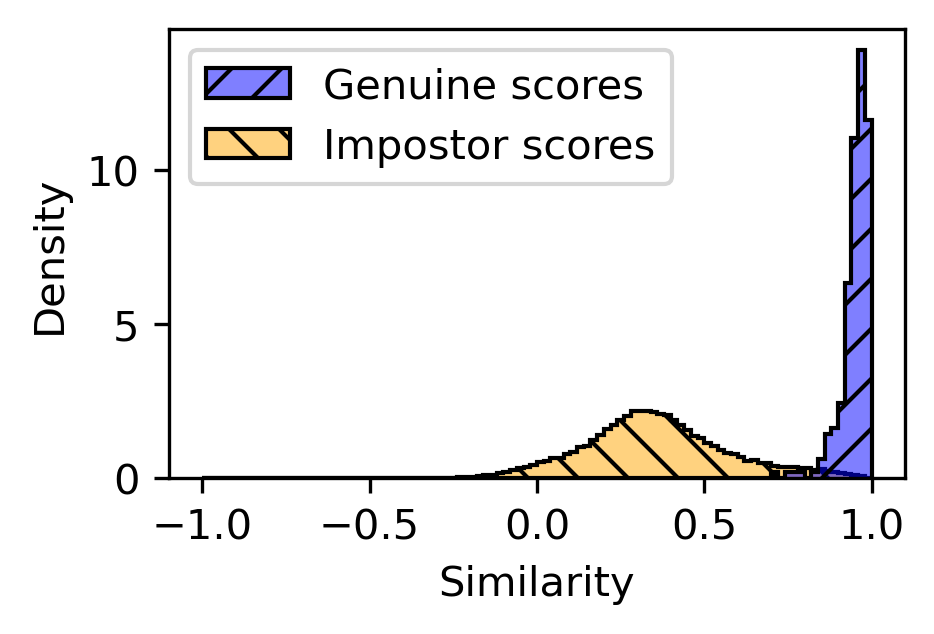}
    \includegraphics[width=0.49\linewidth]{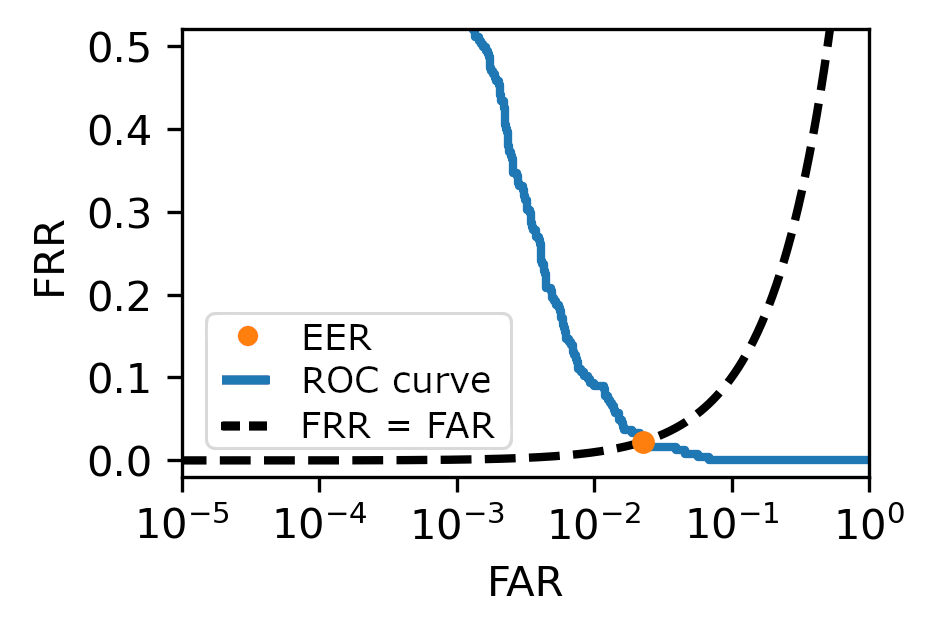} \\
    \caption{Experiment~\expBothBothLow{} using 20~seconds for enrollment and verification.  Left: histograms of the genuine and impostor similarity score distributions.  Right: the ROC curve, with an orange dot indicating the EER and a dashed black line where $\text{FRR} = \text{FAR}$.  The top figures are with the high error test set, the center figures are with the mid error test set, and the bottom figures are with the low error test set.}
    \label{fig:sim-roc-exp71}
\end{figure*}

\begin{figure*}
    \centering
    \includegraphics[width=0.49\linewidth]{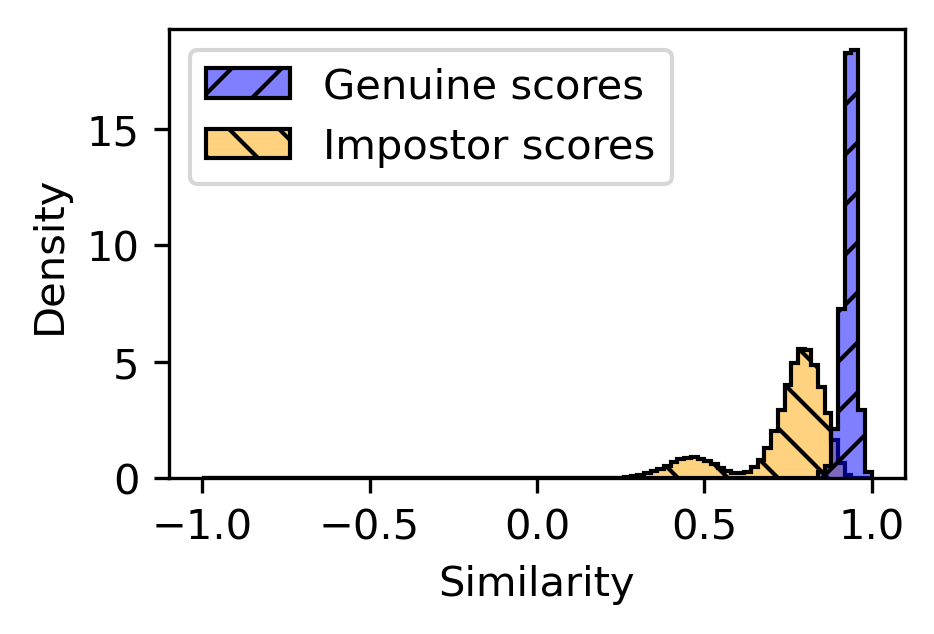}
    \includegraphics[width=0.49\linewidth]{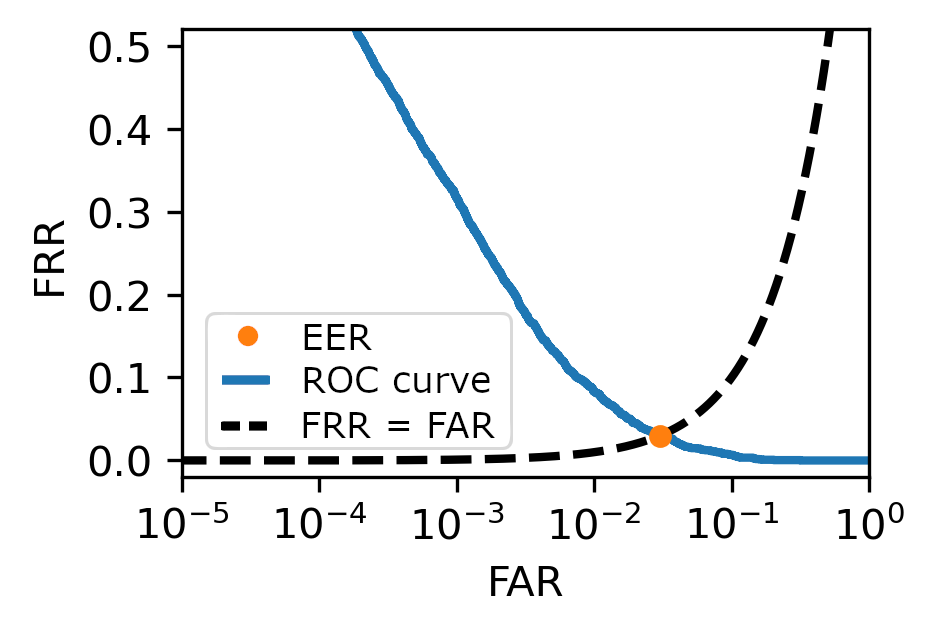}
    \caption{Experiment~\expBothVisualBig{} using 20~seconds for enrollment and verification.  Left: histograms of the genuine and impostor similarity score distributions.  Right: the ROC curve, with an orange dot indicating the EER and a dashed black line where $\text{FRR} = \text{FAR}$.}
    \label{fig:sim-roc-exp53}
\end{figure*}

\begin{figure*}
    \centering
    \includegraphics[width=0.49\linewidth]{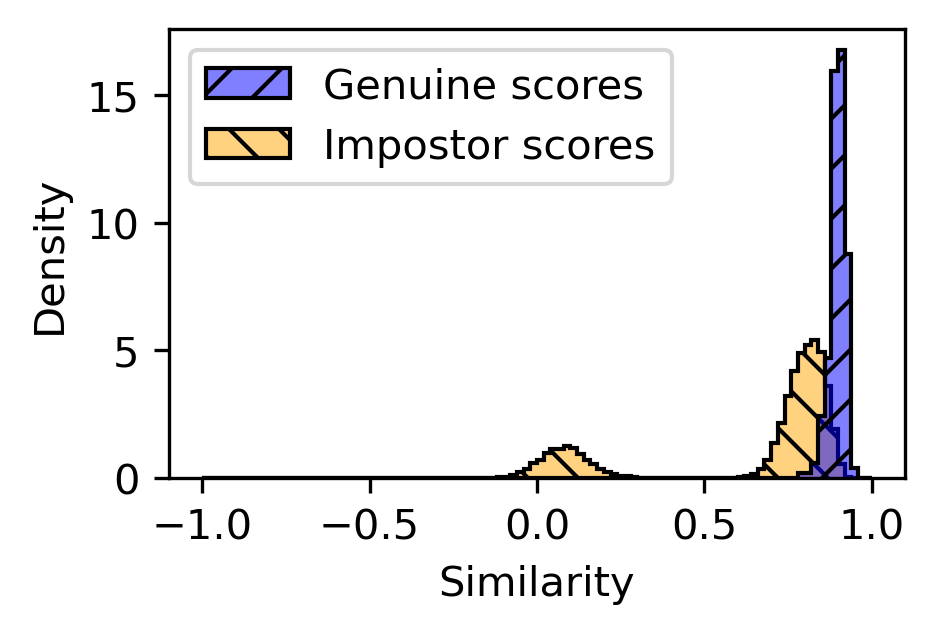}
    \includegraphics[width=0.49\linewidth]{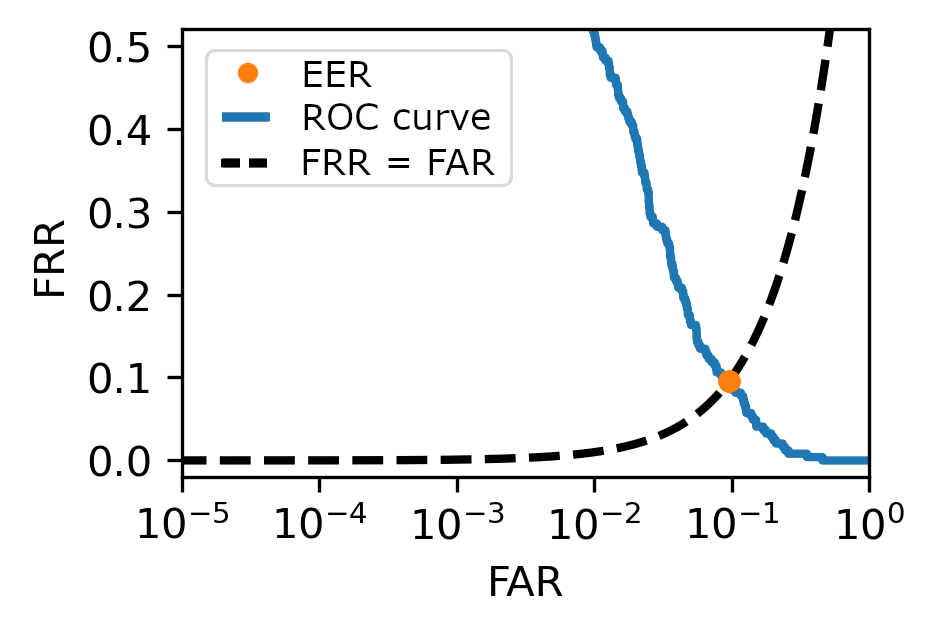} \\
    \includegraphics[width=0.49\linewidth]{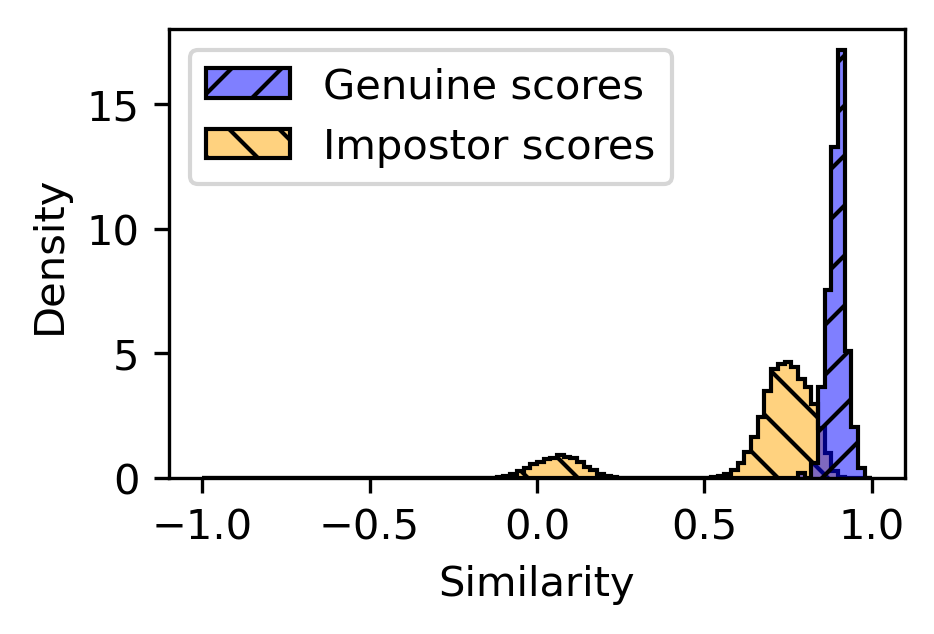}
    \includegraphics[width=0.49\linewidth]{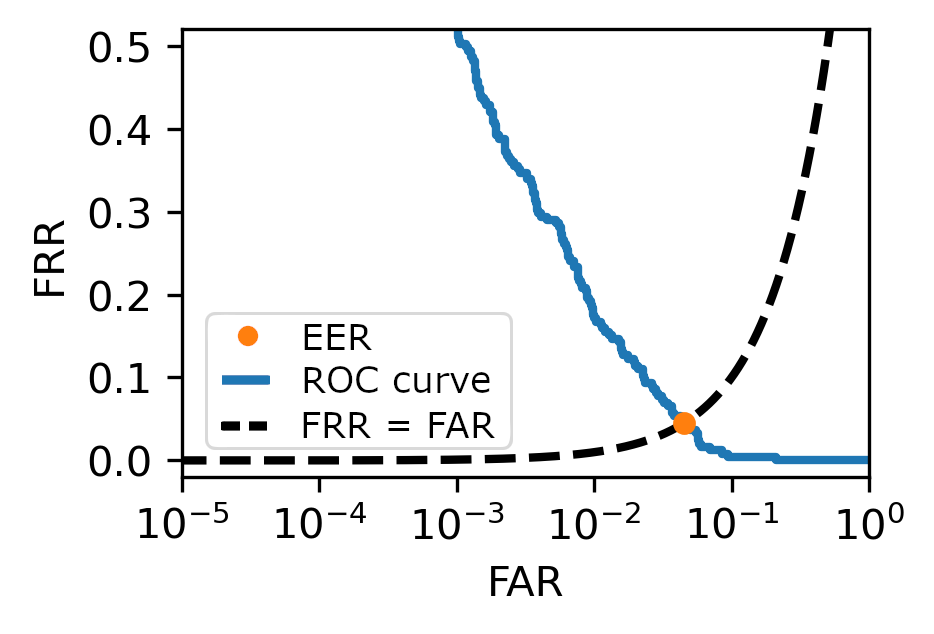} \\
    \includegraphics[width=0.49\linewidth]{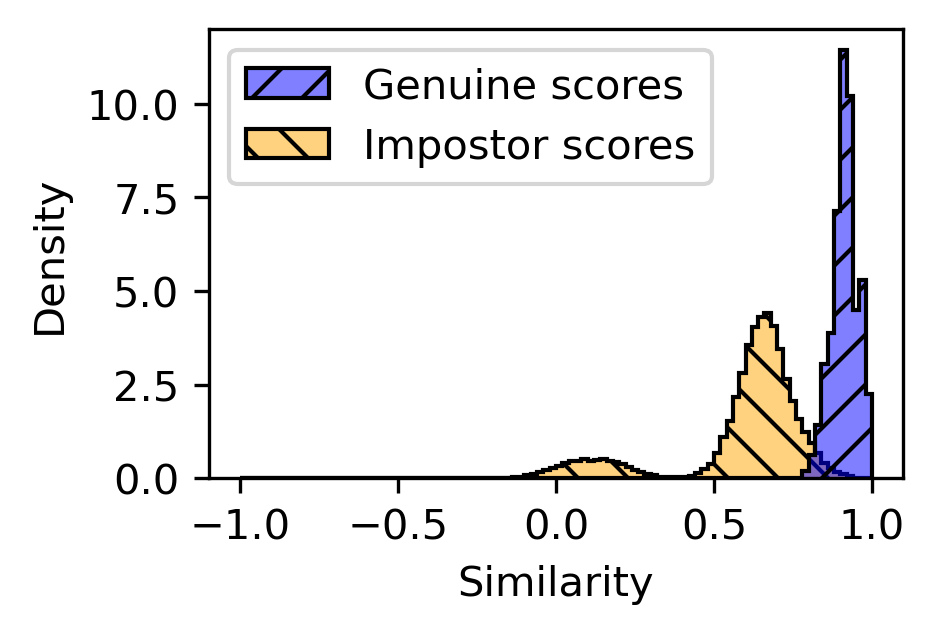}
    \includegraphics[width=0.49\linewidth]{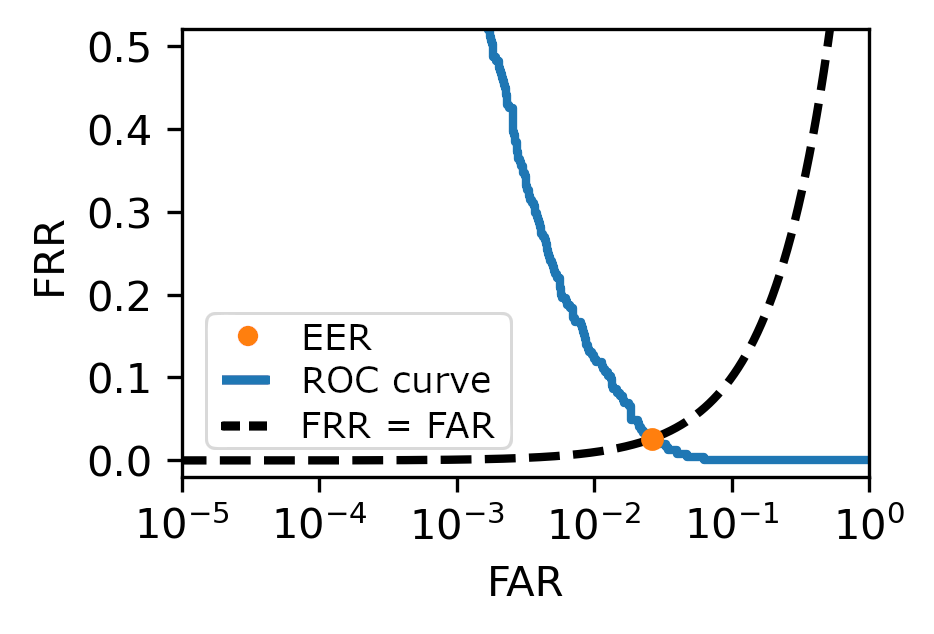} \\
    \caption{Experiment~\expBothVisualHigh{} using 20~seconds for enrollment and verification.  Left: histograms of the genuine and impostor similarity score distributions.  Right: the ROC curve, with an orange dot indicating the EER and a dashed black line where $\text{FRR} = \text{FAR}$.  The top figures are with the high error test set, the center figures are with the mid error test set, and the bottom figures are with the low error test set.}
    \label{fig:sim-roc-exp59}
\end{figure*}

\begin{figure*}
    \centering
    \includegraphics[width=0.49\linewidth]{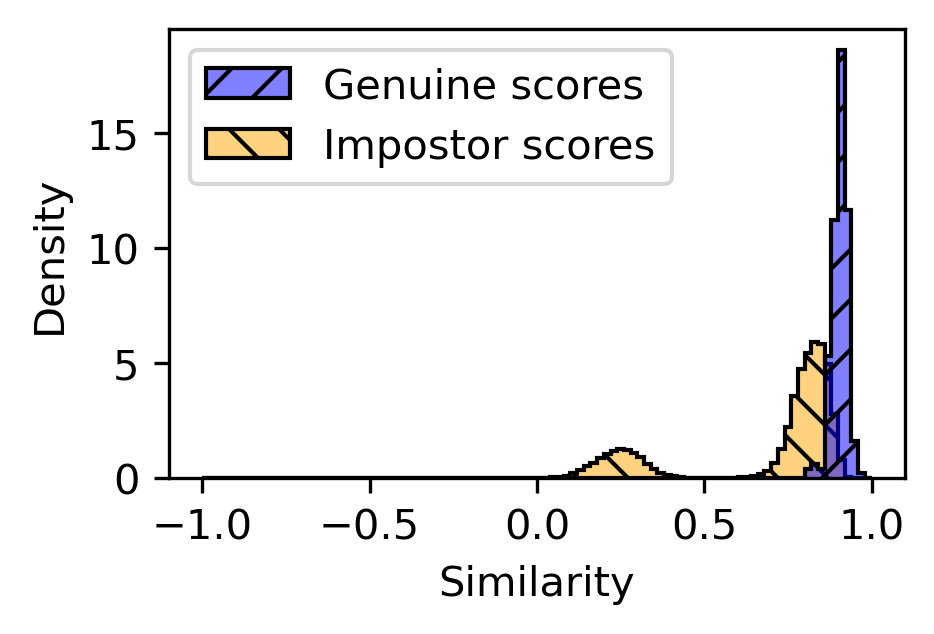}
    \includegraphics[width=0.49\linewidth]{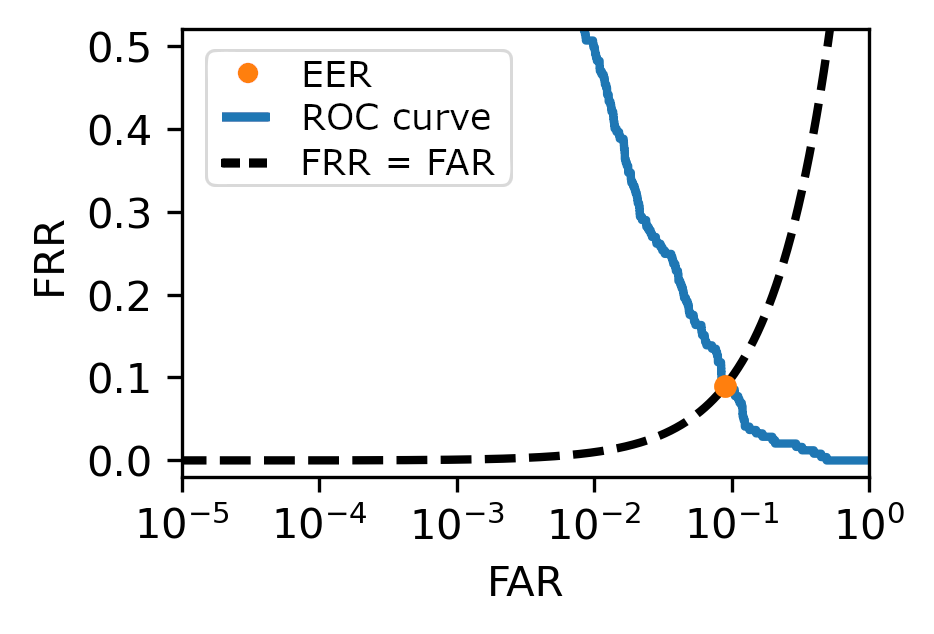} \\
    \includegraphics[width=0.49\linewidth]{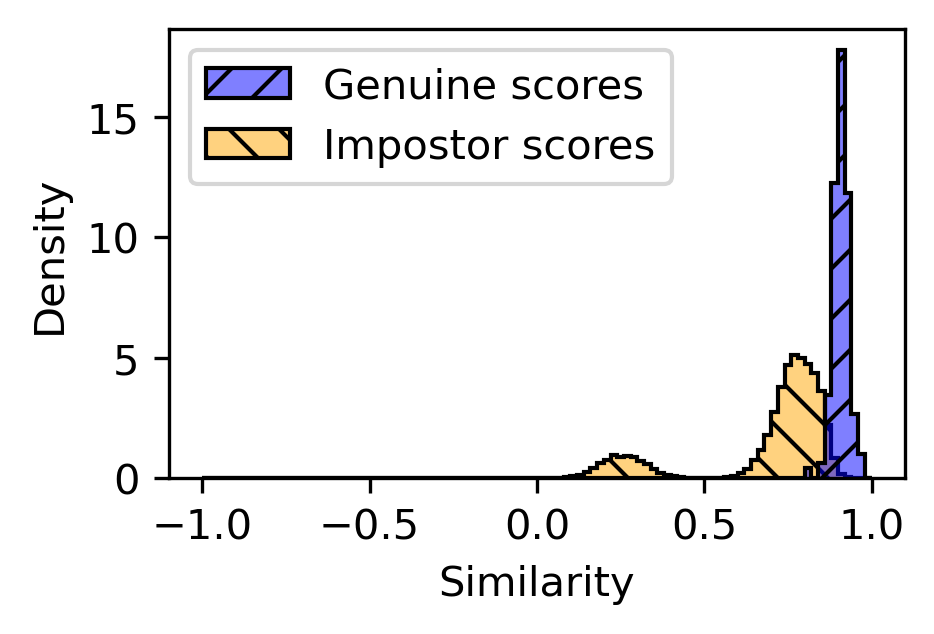}
    \includegraphics[width=0.49\linewidth]{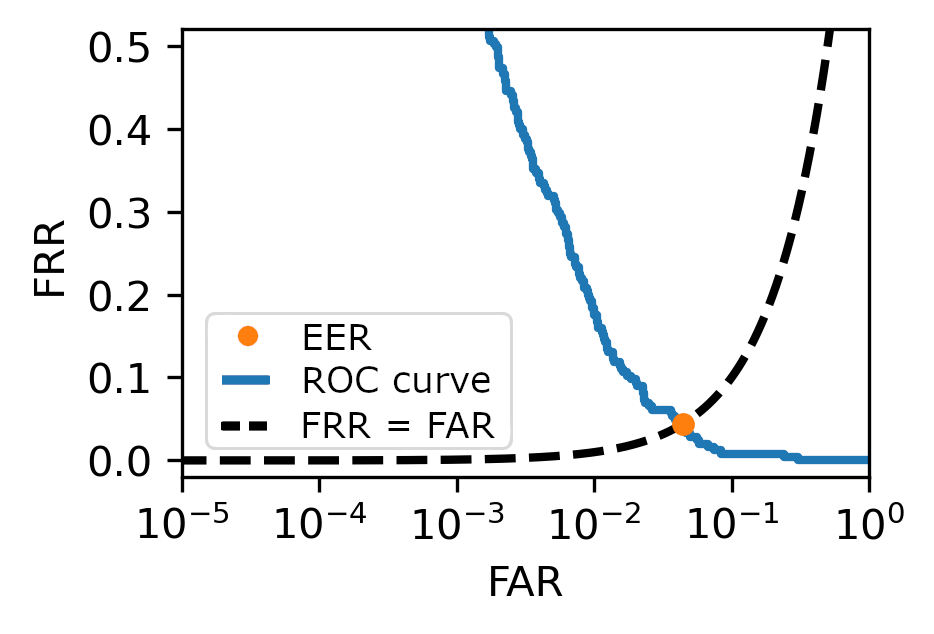} \\
    \includegraphics[width=0.49\linewidth]{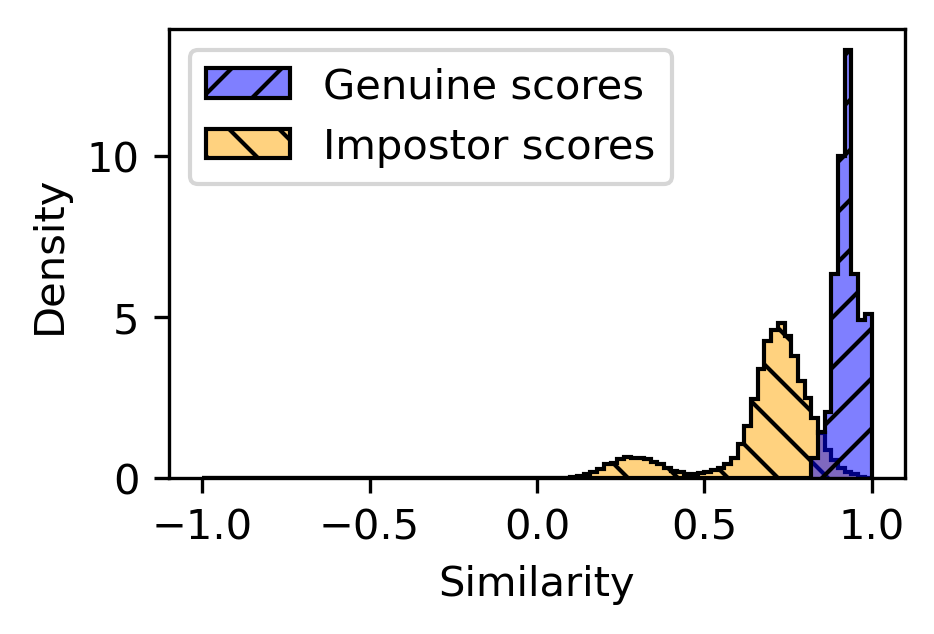}
    \includegraphics[width=0.49\linewidth]{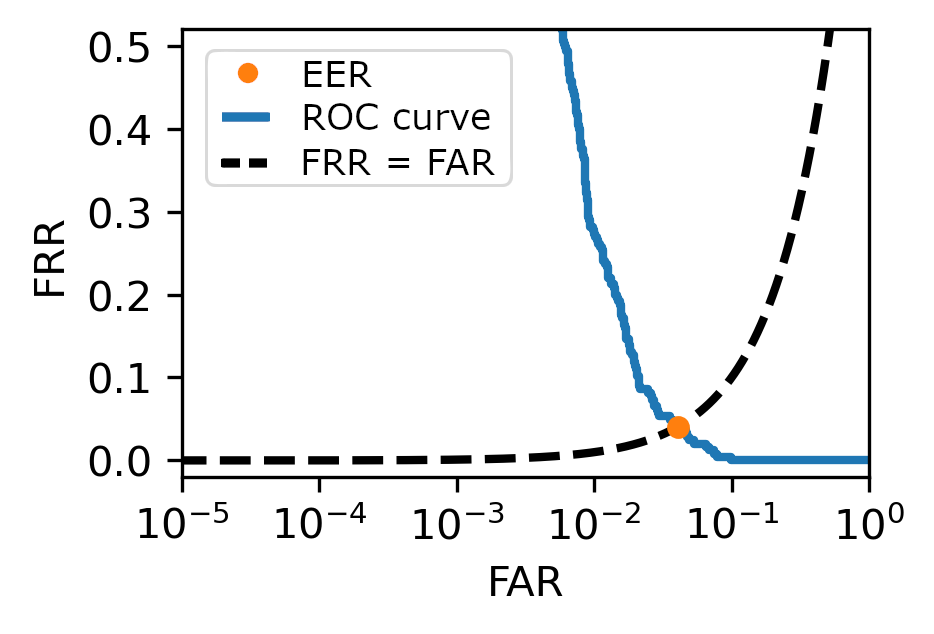} \\
    \caption{Experiment~\expBothVisualLow{} using 20~seconds for enrollment and verification.  Left: histograms of the genuine and impostor similarity score distributions.  Right: the ROC curve, with an orange dot indicating the EER and a dashed black line where $\text{FRR} = \text{FAR}$.  The top figures are with the high error test set, the center figures are with the mid error test set, and the bottom figures are with the low error test set.}
    \label{fig:sim-roc-exp60}
\end{figure*}

From the \gls{roc} curves (as seen in Figures~\ref{fig:sim-roc-exp58}--\ref{fig:sim-roc-exp60}), we are able to estimate several points of interest.
One of these points of interest is the \gls{eer}, which is the point where the \gls{frr} and the \gls{far} are equal.
If there is no point where the two are exactly equal, the point of equality is estimated using linear interpolation.
Another point of interest is the \gls{frr} at a given \gls{far}, denoted $\text{FRR}_{X\%}$ where $X\%$ is the \gls{far} expressed as a percentage.
For example, $\text{FRR}_{0.01\%} = 5\%$ would indicate that the \gls{frr} is \num{5}\% (\num{1}-in-\num{20} odds) when the \gls{far} is \num{0.01}\% (\num{1}-in-\num{10000} odds).
A lower \gls{eer} and a lower $\text{FRR}_{X\%}$ both indicate better performance.

In addition to measures derived from the \gls{roc} curve, we also use a threshold-free measure, the \gls{dprime}~\cite{daugman2000biometric} (also called the sensitivity index or the discriminability index), that represents the degree of separation between the genuine and impostor similarity score distributions.
Given two univariate distributions with possibly unequal variance---in our case, the genuine and impostor similarity score distributions---\gls{dprime} is computed as
\begin{equation}
   d' = \frac{\lvert \mu_1 - \mu_2 \rvert}{\sigma_{\text{rms}}},\, \sigma_{\text{rms}} = \sqrt{\frac{1}{2} \left( \sigma_1^2 + \sigma_2^2 \right)},
\end{equation}
where $\mu_1,\, \mu_2$ are the means of the two distributions and $\sigma_1,\, \sigma_2$ are the \glspl{sd} of the two distributions.
A higher \gls{dprime} indicates a greater degree of separation between the genuine and impostor distributions, and thus better performance.
However, \gls{dprime} becomes less informative as the two distributions deviate from normality and unimodality.

\subsubsection{Identification}
\label{sec:identification}
We follow the same process for identification as for verification to obtain similarity scores and labels for all possible verification attempts.
Users who are in the verification set but not in the enrollment set are excluded from this analysis.
Then, for each remaining user in the verification set, we match that user with the enrolled record of a person with the highest similarity to that user.
A match is correct if the two records represent the same person or incorrect if they represent different people.
We can then compute the rank-1 \gls{ir} as a percentage using
\begin{equation}
    \text{Rank-1 IR} = 100 \cdot \frac{\# \text{correct}}{\# \text{correct} + \# \text{incorrect}}.
\end{equation}

\section{Results}

In Table~\ref{tab:20sec}, we present results for a wide range of experiments when using \num{20}~seconds of data for both enrollment and verification.
In Table~\ref{tab:5sec}, we present the same when using \num{5}~seconds of data.
Note that Experiments~\expBothVisualBig{}--\expBothOptical{} are new to this supplementary material.

\subsection{Testing population size}

\begin{figure*}
    \centering
    \includegraphics[width=0.49\linewidth]{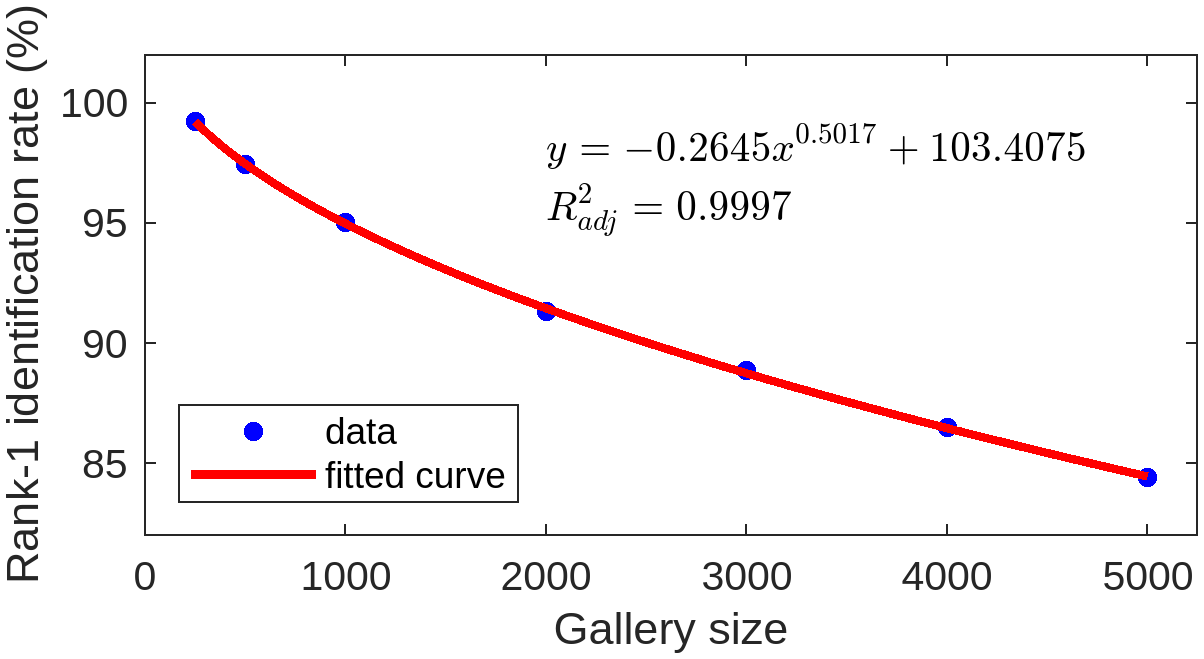}
    \includegraphics[width=0.49\linewidth]{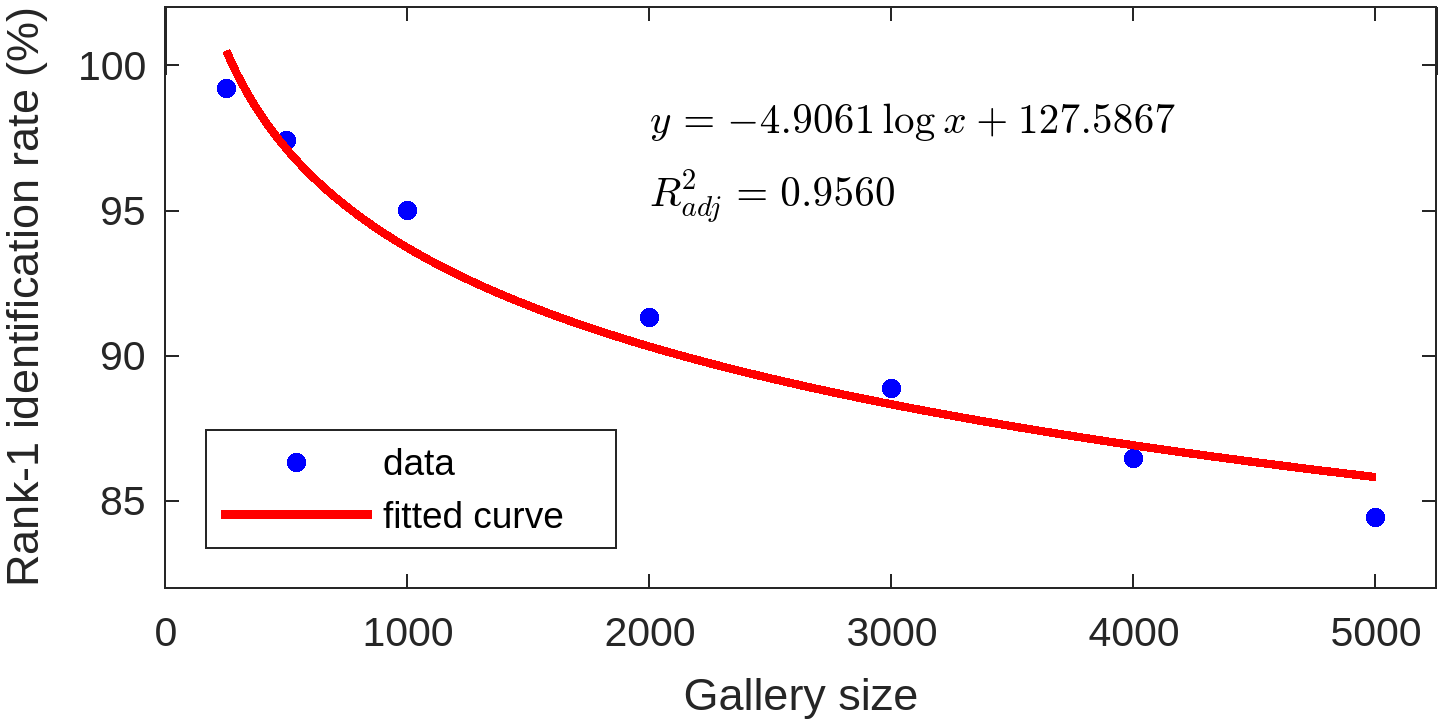} \\
    \includegraphics[width=0.49\linewidth]{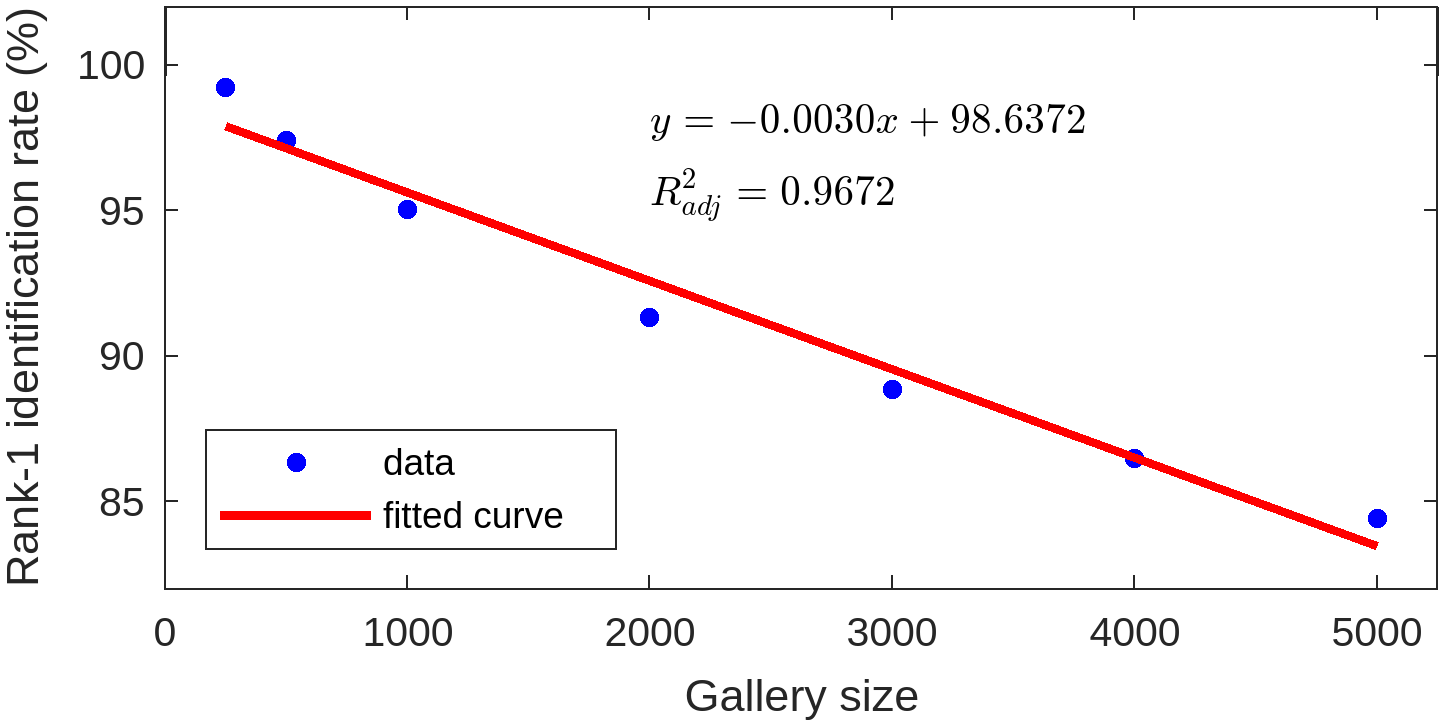}
    \includegraphics[width=0.49\linewidth]{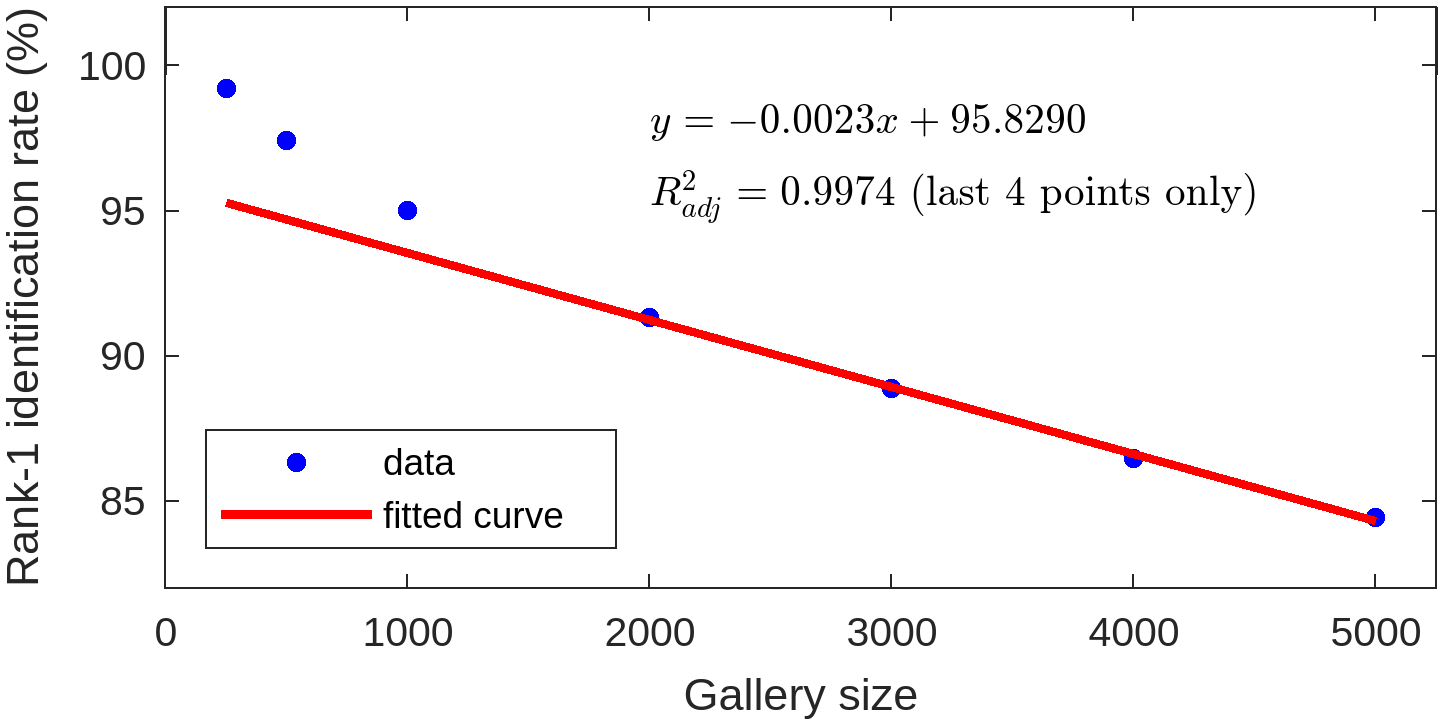}
    \caption{Observations of identification rate (P95 across \num{100}~random samples) vs gallery size.  Different curves are fit in each plot.  By extrapolating from these curves, we estimate the gallery size at which the identification rates might reach below-chance levels.  Top-left: $ax^b+c$ yields an estimate of \num{147000}~users.  Top-right: $a\log{x}+b$ yields an estimate of $2 \times 10^{11}$~users.  Bottom-left: $ax+b$ yields an estimate of \num{32000}~users.  Bottom-right: $ax+b$ (fit only to the last four observations which appear to follow a linear trend) yields an estimate of \num{42000}.}
    \label{fig:fit}
\end{figure*}

In the main manuscript, we fit a curve of the form $a\sqrt{x}+b$ to the observations of identification rate vs gallery size.
This curve was chosen because it had the best fit (measured by $R^2_{adj}$) out of several tested curves.
Here, we present several other possible curves (see Figure~\ref{fig:fit}) and their roots (which serve as our estimate for when identification accuracy might drop below chance levels).

\subsection{Feature permanence}
\begin{figure*}
    \centering
    \includegraphics[width=0.49\linewidth]{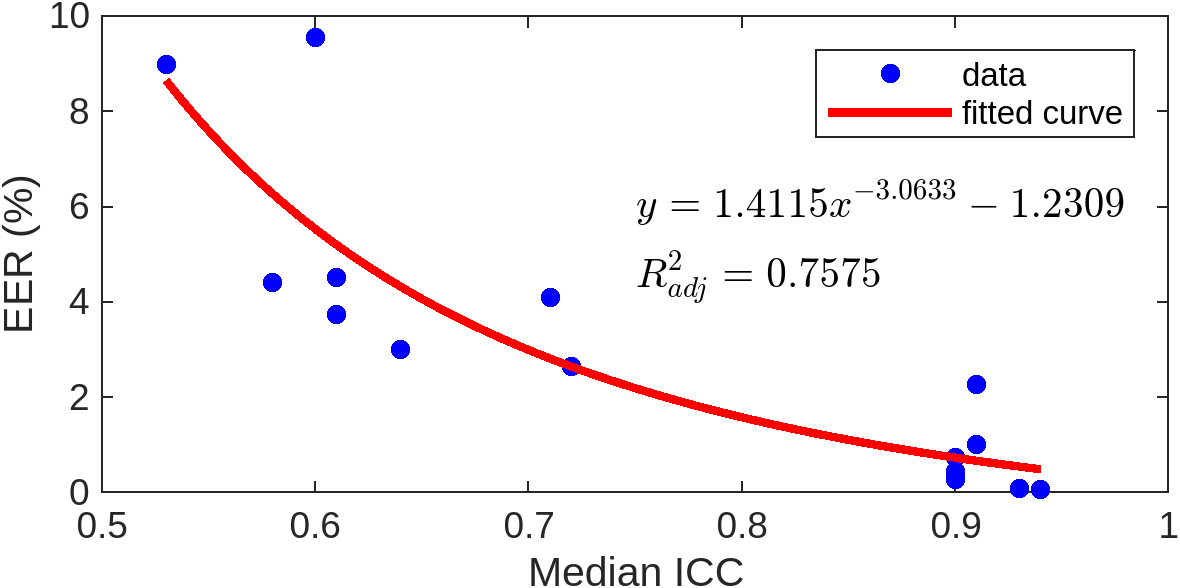}
    \includegraphics[width=0.49\linewidth]{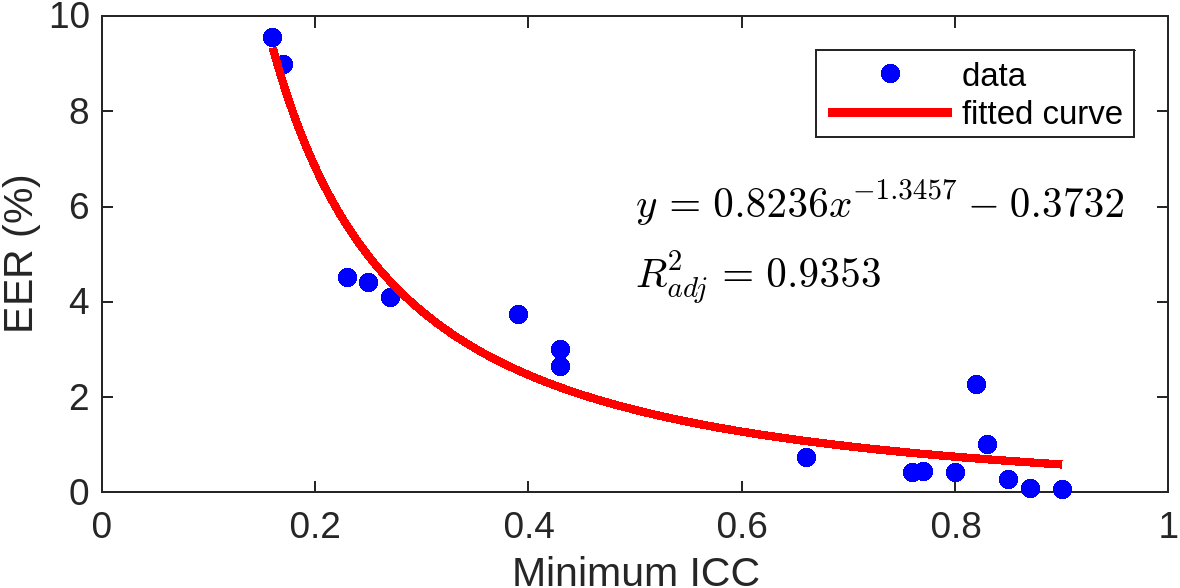}
    \caption{Observations of \gls{eer} vs \gls{icc}.  The left figure takes the median \gls{icc} across the \num{128}~embedding features, while the right figure takes the minimum \gls{icc}.  In each figure, a curve of the form $ax^b+c$ is fit to the observations to highlight the trend.}
    \label{fig:eer-icc}
\end{figure*}

See Table~\ref{tab:permanence} and Figure~\ref{fig:eer-icc}.
To assess normality, we followed the normality test used by Friedman et al.~\cite{friedman_2021_multimodal} comparing the skewness and excess kurtosis of each empirical feature distribution against a population of random normal distributions.
Although the experiments involving different spatial accuracy groups had a greater number of features that passed this normality test, that may be due to those experiments having a smaller gallery size and thus fewer values in their respective feature distributions.

\section{Discussion}

\subsection{Limitations}
The GazePro dataset includes only a single recording session per user, so our study offers a limited assessment of feature permanence; but compared to other studies that include similarly short test-retest intervals, our achieved results are substantially better.
Other studies have conducted longitudinal evaluations and found that a high level of permanence is maintained over intervals as long as \num{3}~years~\cite{lohr_2022_ekyt}.
Our performance measures may be optimistic as a result of the short test-retest interval.
Additionally, the headset may or may not have been removed during a recording session, so there may potentially be no change to headset fitment throughout the session for any given user.

Each user was calibrated only once at the beginning of the recording session, so we had access to only a single set of calibration parameters per user with which visual axis could be estimated.
There is, in essence, only one correct set of calibration parameters for a given user, namely the angle between the visual axis and the optical axis (commonly referred to as \textit{angle kappa}).
An ideal system would be able to perfectly measure angle kappa during calibration, in which case there would be only a single set of calibration parameters per user as there was in our study.
In practice, various factors such as headset fit and user compliance may affect the system's ability to accurately measure the ideal calibration parameters, so different calibration attempts may produce different calibration parameters.

It is important to note that the use of the visual axis assumes that we have access to user calibration parameters and that these parameters are fit prior to every verification attempt.
Our visual axis estimates use each user's own calibration parameters, but there are scenarios such as continuous authentication in which a calibration procedure could not reasonably be performed prior to each verification attempt.
In such scenarios, we would need to either use only the optical axis or estimate the visual axis using the calibration parameters of the enrolled person (as opposed to the calibration parameters of the user attempting verification).
Further investigation is needed to determine authentication accuracy when using the enrolled person's calibration parameters to estimate the visual axis rather than the presenting user's own calibration parameters.

Lastly, we note that our similarity score distributions often deviate from normality and unimodality, so our measures of \gls{dprime} may be less informative.

\bibliographystyle{ieee}
\bibliography{supp}

\clearpage
\begin{table*}
    \centering
    \caption{Self-reported race/ethnicity of participants in the GazePro dataset.}
    \label{tab:race-ethnicity}
    \begin{tabular}{l S[table-format=4.0]}
        \toprule
        Race/ethnicity & {N} \\
        \midrule 
        American Indian or Alaska Native & 53 \\
        Black or African American & 1695 \\
        East Asian & 869 \\
        Hispanic or Latino & 1007 \\
        Middle Eastern & 77 \\
        Native Hawaiian or Other Pacific Islander & 43 \\
        South Asian & 508 \\
        White & 4619 \\
        Other & 331 \\
        \bottomrule
    \end{tabular}
\end{table*}

\clearpage
\begin{table*}
    \centering
    \caption{Verification performance when using \num{20}~seconds of data from random saccade tasks.  Up/down arrows indicate whether higher/lower values are better.  To reduce visual clutter, the symbol ``\ditto'' is used to indicate that the value is repeated from the row above.  In the ``Eye'' column, ``L'' is the left eye and ``R'' is the right eye.  In the ``Axis'' column, ``O'' is the optical axis and ``V'' is the visual axis.}
    \label{tab:20sec}
    \adjustbox{scale=0.8}{
    \begin{tabular}{l l l l l l S[table-format=2.2] S[table-format=3.1] S[table-format=1.1]}
        \toprule
        Experiment & Eye & Axis & Train set & Test set & {EER (\%) $\downarrow$} & {$\text{FRR}_{0.002\%}$ (\%) $\downarrow$} & {$d'$ $\uparrow$} \\
        \midrule
        \expLeftVisual{} & L & V & Full & Full & 8.53 & 99.2 & 1.4 \\
        \expBothVisual{} & L,R	& {\ditto}	& {\ditto}	& {\ditto} & 3.73	& 93.1 & 1.9 \\
        \midrule
        \expBothBoth{} & L,R	& O,V	& Full	& Full & 0.08	& 17.7 & 5.9 \\
        \expBothDiff{} & {\ditto}	& $\text{V} - \text{O}$	& {\ditto}	& {\ditto} & 0.27	& 45.6 & 6.0 \\
        \midrule
        \expBothBothBig{} & L,R & O,V & Full, \num{759} epochs, $m=1024$ & Full & 0.04 & 4.1 & 6.4 \\
        \expBothBothBigFolds{} & {\ditto} & {\ditto} & \num{4} folds, \num{1000} epochs, $m=1024$ & {\ditto} & 0.04 & 2.4 & 6.8 \\
        \midrule
        \expBothBothQuarter{} & L,R	& O,V	& N=\hphantom{0}250	& Full	& 1.59	& 92.2 & 3.7 \\
        \expBothBothHalf{} & {\ditto}	& {\ditto}	& N=\hphantom{0}500	& {\ditto}	& 0.98	& 84.1 & 4.7	\\
        \expBothBothOne{} & {\ditto}	& {\ditto}	& N=1000	& {\ditto}	& 0.66	& 82.6 & 5.2 \\
        \expBothBothTwo{} & {\ditto}	& {\ditto}	& N=2000	& {\ditto}	& 0.61	& 78.9 & 5.4 \\
        \expBothBothThree{} & {\ditto}	& {\ditto}	& N=3000	& {\ditto}	& 0.10	& 22.0 & 5.7 \\
        \expBothBothFour{} & {\ditto}	& {\ditto}	& N=4000	& {\ditto}	& 0.42	& 71.7 & 5.2 \\
        \expBothBothFive{} & {\ditto}	& {\ditto}	& N=5000	& {\ditto}	& 0.07	& 13.7 & 6.1 \\
        \expBothBothSix{} & {\ditto}	& {\ditto}	& N=6000	& {\ditto}	& 0.06	& 12.6 & 6.0 \\
        \midrule
        \expBothBothFoldsFour{}@\num{250} & L,R	& O,V	& \num{4}~folds, N=4000	& N=\hphantom{0}250 & 0.42	& 65.9 & 5.6 \\
        \expBothBothFoldsFour{}@\num{500} & {\ditto}	& {\ditto}	& {\ditto}	& N=\hphantom{0}500 & 0.45	& 68.9 & 5.6 \\
        \expBothBothFoldsFour{}@\num{1}K & {\ditto}	& {\ditto}	& {\ditto}	& N=1000 & 0.41	& 70.4 & 5.6 \\
        \expBothBothFoldsFour{}@\num{2}K & {\ditto}	& {\ditto}	& {\ditto}	& N=2000 & 0.43	& 69.9 & 5.6 \\
        \expBothBothFoldsFour{}@\num{3}K & {\ditto}	& {\ditto}	& {\ditto}	& N=3000 & 0.44	& 71.3 & 5.6 \\
        \expBothBothFoldsFour{}@\num{4}K & {\ditto}	& {\ditto}	& {\ditto}	& N=4000 & 0.44	& 71.0 & 5.6 \\
        \expBothBothFoldsFour{}@\num{5}K & {\ditto}	& {\ditto}	& {\ditto}	& N=5000 & 0.44	& 70.9 & 5.6 \\
        \midrule
        \expBothBothHigh{}@HighErr & L,R	& O,V	& High err.	& High err.	& 1.00	& 75.9 & 5.0 \\
        \expBothBothHigh{}@MidErr & {\ditto}	& {\ditto}	& {\ditto}	& Mid err. & 0.41	& 75.9 & 5.1 \\
        \expBothBothHigh{}@LowErr & {\ditto}	& {\ditto}	& {\ditto}	& Low err. & 0.73	& 76.7 & 4.2 \\
        \expBothBothLow{}@HighErr & {\ditto}	& {\ditto}	& Low err.	& High err. & 2.25	& 91.8 & 3.9 \\
        \expBothBothLow{}@MidErr & {\ditto}	& {\ditto}	& {\ditto}	& Mid err.	& 0.41	& 70.6 & 4.8 \\
        \expBothBothLow{}@LowErr & {\ditto}	& {\ditto}	& {\ditto}	& Low err.	& 0.42	& 92.2 & 4.4 \\
        \midrule
        \expBothVisualBig{} & L,R & V & Full, \num{1000} epochs, $m=1024$ & Full & 2.99 & 76.7 & 2.0 \\
        \midrule
        \expBothVisualQuarter{} & L,R	& V	& N=250	& Full	& 19.16	& 99.7 & 1.6 \\
        \expBothVisualHalf{} & {\ditto}	& {\ditto}	& N=500	& {\ditto}	& 15.86	& 99.4 & 1.6 \\
        \expBothVisualOne{} & {\ditto}	& {\ditto}	& N=1000	& {\ditto}	& 6.23	& 96.7 & 2.0 \\
        \expBothVisualTwo{} & {\ditto}	& {\ditto}	& N=2000	& {\ditto}	& 4.96	& 95.3 & 1.7 \\
        \expBothVisualThree{} & {\ditto}	& {\ditto}	& N=3000	& {\ditto}	& 4.20	& 93.2 & 1.9 \\
        \expBothVisualFour{} & {\ditto}	& {\ditto}	& N=4000	& {\ditto}	& 4.18	& 94.2 & 2.0 \\
        \expBothVisualFive{} & {\ditto}	& {\ditto}	& N=5000	& {\ditto}	& 4.04	& 94.2 & 1.8 \\
        \expBothVisualSix{} & {\ditto}	& {\ditto}	& N=6000	& {\ditto}	& 4.07	& 92.2 & 1.8 \\
        \midrule
        \expBothVisualHigh{}@HighErr & L,R	& V	& High err.	& High err.	& 2.63	& 87.4 & 2.2 \\
        \expBothVisualHigh{}@MidErr & {\ditto}	& {\ditto}	& {\ditto}	& Med. err.	& 4.49	& 92.2 & 1.4 \\
        \expBothVisualHigh{}@LowErr & {\ditto}	& {\ditto}	& {\ditto}	& Low err.	& 9.53	& 98.8 & 1.2 \\
        \expBothVisualLow{}@HighErr & {\ditto}	& {\ditto}	& Low err.	& High err.	& 4.08	& 94.7 & 2.1 \\
        \expBothVisualLow{}@MidErr & {\ditto}	& {\ditto}	& {\ditto}	& Med. err.	& 4.39	& 96.7 & 1.5 \\
        \expBothVisualLow{}@LowErr & {\ditto}	& {\ditto}	& {\ditto}	& Low err.	& 8.98	& 94.3 & 1.2 \\
        \midrule
        \expLeftBoth{} & L & O,V & Full & Full & 0.38	& 71.3 & 5.3 \\
        \expBothOptical{} & L,R	& O	& {\ditto}	& {\ditto}	& 7.93	& 98.5 & 1.4 \\
        \bottomrule
    \end{tabular}
    }
\end{table*}

\clearpage
\begin{table*}
    \centering
    \caption{Verification performance when using \num{5}~seconds of data from random saccade tasks.  Up/down arrows indicate whether higher/lower values are better.  To reduce visual clutter, the symbol ``\ditto'' is used to indicate that the value is repeated from the row above.  In the ``Eye'' column, ``L'' is the left eye and ``R'' is the right eye.  In the ``Axis'' column, ``O'' is the optical axis and ``V'' is the visual axis.}
    \label{tab:5sec}
    \adjustbox{scale=0.8}{
    \begin{tabular}{l l l l l l S[table-format=2.2] S[table-format=3.1] S[table-format=1.1]}
        \toprule
        Experiment & Eye & Axis & Train set & Test set & {EER (\%) $\downarrow$} & {$\text{FRR}_{0.002\%}$ (\%) $\downarrow$} & {$d'$ $\uparrow$} \\
        \midrule
        \expLeftVisual{} & L & V & Full & Full & 24.21 & 99.5 & 1.2 \\
        \expBothVisual{} & L,R	& {\ditto}	& {\ditto}	& {\ditto} & 13.14	& 97.3 & 1.8 \\
        \midrule
        \expBothBoth{} & L,R	& O,V	& Full	& Full & 0.46	& 72.0 & 5.4 \\
        \expBothDiff{} & {\ditto}	& $\text{V} - \text{O}$	& {\ditto}	& {\ditto} & 1.23	& 83.8 & 4.9 \\
        \midrule
        \expBothBothBig{} & L,R & O,V & Full, \num{759} epochs, $m=1024$ & Full & 0.18 & 41.9 & 6.0 \\
        \expBothBothBigFolds{} & {\ditto} & {\ditto} & \num{4} folds, \num{1000} epochs, $m=1024$ & {\ditto} & 0.17 & 32.8 & 6.4 \\
        \midrule
        \expBothBothQuarter{} & L,R	& O,V	& N=\hphantom{0}250	& Full	& 4.22	& 95.9 & 3.3 \\
        \expBothBothHalf{} & {\ditto}	& {\ditto}	& N=\hphantom{0}500	& {\ditto}	& 2.75	& 93.9 & 4.0	\\
        \expBothBothOne{} & {\ditto}	& {\ditto}	& N=1000	& {\ditto}	& 2.25	& 93.6 & 4.3 \\
        \expBothBothTwo{} & {\ditto}	& {\ditto}	& N=2000	& {\ditto}	& 1.82	& 92.6 & 4.5 \\
        \expBothBothThree{} & {\ditto}	& {\ditto}	& N=3000	& {\ditto}	& 0.48	& 74.1 & 5.3 \\
        \expBothBothFour{} & {\ditto}	& {\ditto}	& N=4000	& {\ditto}	& 1.48	& 91.1 & 4.4 \\
        \expBothBothFive{} & {\ditto}	& {\ditto}	& N=5000	& {\ditto}	& 0.33	& 68.0 & 5.6 \\
        \expBothBothSix{} & {\ditto}	& {\ditto}	& N=6000	& {\ditto}	& 0.38	& 67.7 & 5.5 \\
        \midrule
        \expBothBothFoldsFour{}@\num{250} & L,R	& O,V	& \num{4}~folds, N=4000	& N=\hphantom{0}250 & 1.24	& 89.0 & 4.9 \\
        \expBothBothFoldsFour{}@\num{500} & {\ditto}	& {\ditto}	& {\ditto}	& N=\hphantom{0}500 & 1.28	& 90.6 & 4.9 \\
        \expBothBothFoldsFour{}@\num{1}K & {\ditto}	& {\ditto}	& {\ditto}	& N=1000 & 1.23	& 91.2 & 4.9 \\
        \expBothBothFoldsFour{}@\num{2}K & {\ditto}	& {\ditto}	& {\ditto}	& N=2000 & 1.21	& 90.9 & 4.9 \\
        \expBothBothFoldsFour{}@\num{3}K & {\ditto}	& {\ditto}	& {\ditto}	& N=3000 & 1.21	& 90.9 & 4.9 \\
        \midrule
        \expBothBothHigh{}@HighErr & L,R	& O,V	& High err.	& High err.	& 2.22	& 86.5 & 4.4 \\
        \expBothBothHigh{}@MidErr & {\ditto}	& {\ditto}	& {\ditto}	& Mid err. & 1.24	& 91.4 & 4.4 \\
        \expBothBothHigh{}@LowErr & {\ditto}	& {\ditto}	& {\ditto}	& Low err. & 2.04	& 91.8 & 3.9 \\
        \expBothBothLow{}@HighErr & {\ditto}	& {\ditto}	& Low err.	& High err. & 4.90	& 96.7 & 3.5 \\
        \expBothBothLow{}@MidErr & {\ditto}	& {\ditto}	& {\ditto}	& Mid err.	& 1.63	& 91.4 & 4.1 \\
        \expBothBothLow{}@LowErr & {\ditto}	& {\ditto}	& {\ditto}	& Low err.	& 2.04	& 94.7 & 3.8 \\
        \midrule
        \expBothVisualBig{} & L,R & V & Full, \num{1000} epochs, $m=1024$ & Full & 10.86 & 96.5 & 1.8 \\
        \midrule
        \expBothVisualQuarter{} & L,R	& V	& N=250	& Full	& 29.51	& 99.8 & 1.1 \\
        \expBothVisualHalf{} & {\ditto}	& {\ditto}	& N=500	& {\ditto}	& 27.72	& 99.4 & 1.2 \\
        \expBothVisualOne{} & {\ditto}	& {\ditto}	& N=1000	& {\ditto}	& 17.13	& 98.1 & 1.8 \\
        \expBothVisualTwo{} & {\ditto}	& {\ditto}	& N=2000	& {\ditto}	& 13.82	& 97.3 & 1.7 \\
        \expBothVisualThree{} & {\ditto}	& {\ditto}	& N=3000	& {\ditto}	& 13.95	& 97.3 & 1.8 \\
        \expBothVisualFour{} & {\ditto}	& {\ditto}	& N=4000	& {\ditto}	& 14.16	& 97.7 & 1.8 \\
        \expBothVisualFive{} & {\ditto}	& {\ditto}	& N=5000	& {\ditto}	& 13.35	& 97.0 & 1.7 \\
        \expBothVisualSix{} & {\ditto}	& {\ditto}	& N=6000	& {\ditto}	& 13.48	& 96.0 & 1.8 \\
        \midrule
        \expBothVisualHigh{}@HighErr & L,R	& V	& High err.	& High err.	& 8.03	& 96.3 & 2.0 \\
        \expBothVisualHigh{}@MidErr & {\ditto}	& {\ditto}	& {\ditto}	& Med. err.	& 15.92	& 96.7 & 1.4 \\
        \expBothVisualHigh{}@LowErr & {\ditto}	& {\ditto}	& {\ditto}	& Low err.	& 21.19	& 100.0 & 1.2 \\
        \expBothVisualLow{}@HighErr & {\ditto}	& {\ditto}	& Low err.	& High err.	& 11.02	& 95.9 & 1.8 \\
        \expBothVisualLow{}@MidErr & {\ditto}	& {\ditto}	& {\ditto}	& Med. err.	& 15.87	& 98.8 & 1.4 \\
        \expBothVisualLow{}@LowErr & {\ditto}	& {\ditto}	& {\ditto}	& Low err.	& 20.82	& 98.0 & 1.3 \\
        \midrule
        \expLeftBoth{} & L & O,V & Full & Full & 1.31	& 91.6 & 4.6 \\
        \expBothOptical{} & L,R	& O	& {\ditto}	& {\ditto}	& 22.21	& 98.6 & 1.3 \\
        \bottomrule
    \end{tabular}
    }
\end{table*}

\clearpage
\begin{table*}
    \centering
    \caption{Assessment of normality, reliability, and intercorrelations of learned embedding features for different experiments.  The minimum intercorrelation was $0.00$ for all experiments, so it is excluded from this table.}
    \label{tab:permanence}
    \begin{tabular}{@{} lS[table-format=2] S[table-format=1.2]S[table-format=1.2]S[table-format=1.2] S[table-format=1.2]S[table-format=1.2] S[table-format=-1.2]S[table-format=-1.2]S[table-format=1.2] S[table-format=-1.2]S[table-format=-1.2]S[table-format=1.2] @{}}
        \toprule
        {} & {} & \multicolumn{3}{c}{ICC} & \multicolumn{2}{c}{Intercorr.} & \multicolumn{3}{c}{Skewness} & \multicolumn{3}{c}{Excess kurtosis} \\ \cmidrule(lr){3-5} \cmidrule(lr){6-7} \cmidrule(lr){8-10} \cmidrule(lr){11-13}
        {Experiment} & {Normal} & {min} & {med} & {max} & {med} & {max} & {min} & {med} & {max} & {min} & {med} & {max} \\ 
        \midrule
        \expBothVisual{} & 29 & 0.39 & 0.61 & 0.95 & 0.14 & 0.66 & -1.27 & -0.02 & 1.23 & -0.34 & 0.06 & 2.43 \\
        \expBothBoth{} & 17 & 0.87 & 0.93 & 0.99 & 0.10 & 0.54 & -0.73 & 0.02 & 0.85 & -0.72 & -0.15 & 0.70 \\
        \expBothDiff{} & 18 & 0.85 & 0.90 & 0.97 & 0.08 & 0.49 & -0.38 & 0.01 & 1.14 & -0.42 & -0.10 & 1.32 \\
        \expBothBothBig{} & 30 & 0.90 & 0.94 & 0.98 & 0.08 & 0.57 & -0.44 & -0.03 & 1.56 & -0.56 & -0.10 & 1.85 \\
        \expBothBothHigh{}@HighErr & 59 & 0.83 & 0.91 & 0.98 & 0.12 & 0.58 & -0.81 & -0.06 & 0.67 & -0.71 & -0.18 & 1.71 \\
        \expBothBothHigh{}@MidErr & 70 & 0.76 & 0.90 & 0.99 & 0.12 & 0.58 & -1.11 & -0.02 & 0.80 & -0.67 & -0.18 & 1.52 \\
        \expBothBothHigh{}@LowErr & 69 & 0.66 & 0.90 & 0.99 & 0.14 & 0.73 & -1.31 & -0.01 & 0.67 & -0.68 & -0.18 & 1.68 \\
        \expBothBothLow{}@HighErr & 41 & 0.82 & 0.91 & 0.99 & 0.16 & 0.67 & -1.76 & -0.01 & 0.70 & -1.10 & -0.13 & 3.37 \\
        \expBothBothLow{}@MidErr & 67 & 0.80 & 0.90 & 0.99 & 0.13 & 0.64 & -1.82 & -0.01 & 0.74 & -0.80 & -0.24 & 2.82 \\
        \expBothBothLow{}@LowErr & 54 & 0.77 & 0.90 & 0.99 & 0.14 & 0.71 & -1.82 & -0.01 & 0.89 & -0.78 & -0.25 & 2.11 \\
        \expBothVisualBig{} & 25 & 0.43 & 0.64 & 0.96 & 0.12 & 0.74 & -1.34 & 0.05 & 1.61 & -0.28 & 0.05 & 3.66 \\
        \expBothVisualHigh{}@HighErr & 54 & 0.43 & 0.72 & 0.97 & 0.14 & 0.69 & -1.78 & 0.02 & 1.22 & -0.70 & 0.05 & 3.33 \\
        \expBothVisualHigh{}@MidErr & 54 & 0.23 & 0.61 & 0.98 & 0.17 & 0.81 & -2.43 & 0.01 & 1.67 & -0.50 & 0.23 & 4.86 \\
        \expBothVisualHigh{}@LowErr & 45 & 0.16 & 0.60 & 0.98 & 0.24 & 0.89 & -2.13 & 0.02 & 1.63 & -0.53 & 0.27 & 2.91 \\
        \expBothVisualLow{}@HighErr & 48 & 0.27 & 0.71 & 0.97 & 0.15 & 0.79 & -2.35 & -0.02 & 1.19 & -0.66 & 0.14 & 5.19 \\
        \expBothVisualLow{}@MidErr & 55 & 0.25 & 0.58 & 0.98 & 0.16 & 0.83 & -2.61 & 0.02 & 1.68 & -0.64 & 0.17 & 5.66 \\
        \expBothVisualLow{}@LowErr & 50 & 0.17 & 0.53 & 0.98 & 0.21 & 0.88 & -2.20 & 0.00 & 1.69 & -0.56 & 0.26 & 3.33 \\
        \bottomrule
    \end{tabular}
\end{table*}

\clearpage

\clearpage

\clearpage

\clearpage

\clearpage

\clearpage

\clearpage

\clearpage

\clearpage

\clearpage

\clearpage

\clearpage

\clearpage
